\documentclass[pdflatex,sn-mathphys-num]{sn-jnl}

\usepackage{graphicx}%
\usepackage{multirow}%
\usepackage{amsmath,amssymb,amsfonts}%
\usepackage{amsthm}%
\usepackage{mathrsfs}%
\usepackage[title]{appendix}%
\usepackage{xcolor}%
\usepackage{textcomp}%
\usepackage{manyfoot}%
\usepackage{booktabs}%
\usepackage{listings}%

\usepackage[american]{babel}

\usepackage{mathtools} 
\usepackage{booktabs} 
\usepackage{tikz} 
\usepackage{amsthm}
\usepackage{algorithm}
\usepackage{algorithmic}
\usepackage{listings}
\usepackage{xcolor}

\theoremstyle{plain}
\newtheorem{theorem}{Theorem}[section]
\newtheorem{proposition}[theorem]{Proposition}

\theoremstyle{definition}

\theoremstyle{remark}

\usepackage{amsmath}
\usepackage{amssymb}
\usepackage{mathtools}
\usepackage{amsthm}

\renewcommand{\k}{k}
\renewcommand{\P}{\mathcal{P}}

\newcommand{\sdim}{d}
\newcommand{\x}{\mathbf{x}}
\newcommand{\y}{\mathbf{y}}
\newcommand{\X}{\mathbf{X}}
\newcommand{\A}{\mathbf{A}}
\newcommand{\B}{\mathbf{B}}
\newcommand{\Z}{\mathbf{Z}}
\newcommand{\I}{\mathcal{I}}
\newcommand{\K}{\mathbf{K}}
\newcommand{\zero}{\mathbf{0}}
\newcommand{\indep}{\perp\!\!\!\perp}
\newcommand{\Ktall}{\mathbf{T}}
\newcommand{\Knew}{\mathbf{D}}

\begin{document}

\title[Bilateral Uncertainty in Additive Bayesian Optimization]{On Thompson Sampling and Bilateral Uncertainty in Additive Bayesian Optimization}


\author*[1]{\fnm{Nathan} \sur{Wycoff}}\email{nwycoff@umass.edu}

\affil*[1]{\orgdiv{Department of Mathematics and Statistics}, \orgname{University of Massachusetts Amherst}, \orgaddress{\street{740 N Pleasant St}, \city{Amherst}, \postcode{00103}, \state{MA}, \country{United States}}}


\abstract{
In Bayesian Optimization (BO), additive assumptions can mitigate the twin difficulties of modeling and searching a complex function in high dimension.
However, common acquisition functions, like the Additive Lower Confidence Bound, ignore pairwise covariances between dimensions, which we'll call \textit{bilateral uncertainty} (BU), imposing a second layer of approximations.
While theoretical results indicate that asymptotically not much is lost in doing so, little is known about the practical effects of this assumption in small budgets.
In this article, we show that by exploiting conditional independence, Thompson Sampling respecting BU can be efficiently conducted.
We use this fact to execute an empirical investigation into the loss incurred by ignoring BU, finding that the additive approximation to Thompson Sampling does indeed have, on balance, worse performance than the exact method, but that this difference is of little practical significance.
This buttresses the theoretical understanding and suggests that the BU-ignoring approximation is sufficient for BO in practice, even in the non-asymptotic regime.
}




\maketitle

\section{Introduction and Motivation}
Bayesian Optimization \citep[BO]{garnett_bayesoptbook_2023} is a conceptually elegant and practically effective means of finding the optimal setting of a complicated system.
Primarily built on Gaussian process (GP) surrogates, it allows for sufficient flexibility to capture a variety of complex dynamics.
However, BO struggles on high dimensional systems for two primary reasons.
First, the very flexibility which is BO's main asset is also a liability in the high dimensional setting, as the number of functions compatible with a given finite dataset explodes with dimension.
This makes it difficult to accurately model the high dimensional function, and leaves many possibilities in relatively unexplored parts of the input space.
These many possibilities are responsible for the second problem with high dimensional BO: the proliferation of local optima of the acquisition functions used to select which points to evaluate the simulator at.

One avenue for addressing these issues is Additive Bayesian Optimization \citep{kandasamy2015high}, which involves using an additive Gaussian process \citep{durrande2011additive,duvenaud2011additive} as the surrogate model.
This entails assuming that the black-box objective is not free to vary as a function of all parameters jointly.
Rather, the function is assumed to be expressible as a sum of lower dimensional functions, each depending on only a few input parameters.
This allows all the input variables to be taken into account while ruling out many of the most complex interactions implicitly allowed by a standard GP kernel, and thus makes progress on the modeling issue so long as the assumptions are not blatantly violated.

It's quite tempting to think that this also solves the acquisition issue.
Say that we have some function $f$ which may be written as a sum of univariate functions: $f(\x) = f_1(x_1) + f_2(x_2) + \ldots + f_P(x_P)$. 
If we knew the functions $f_1,f_2,\ldots,f_P$, we would be able to find the global minimum of $f$ merely by minding the global minimum of each $f_{i}$ and concatenating them. 
This is attractive because, in the context of acquisition functions for BO, we are typically unwilling to make assumptions like convexity or unimodality which would allow for efficient optimization.
Rather, we expect acquisition functions to be nonconvex and multimodal, which is more in line with a Lipschitz assumption.
But guaranteed global optimization of a Lipschitz function over a bounded domain is exponentially difficult in dimension \citep[Chapter 1]{nesterov2013introductory}; in the worst case we cannot do better than a grid search, which is infeasible in large dimension.
So, in high dimension, additivity turns a hopeless problem into a trivial one and allows us to realistically globally optimize $f$ even in very high dimensions.

However, in the black-box setting, we typically do not observe $f_1(x_1),f_2(x_2),\ldots,f_P(x_P)$; we just observe their sum $f(\x)$.
And if we build a posterior distribution on $[f_1(x_1), f_2(x_2),\ldots,f_P(x_P)]$ given $f(\x)$, there must be significant correlation between them, because an increase in one must be balanced out by a decrease on average of the others, whatever their exact values might be. 
This \textit{bilateral uncertainty} manifests itself in the fact that though the predictive mean is an additive function, the predictive variance is a sum of overlapping two dimensional functions:
\begin{equation}\label{eq:var}
    \sigma^2(\x) = \sum_{i=1}^P \sum_{j=1}^P \sigma^2_{i,j}(x_i,x_j) \,.
\end{equation}
So if we wanted to find the point with the highest predictive variance, for example, we would want to maximize \ref{eq:var}.
While not additive because the indices overlap, its form as the sum of two dimensional functions suggests that it might yet be more tractable than the general Lipschitz case.
Unfortunately, this is not so, which can be seen by noting that indefinite quadratic forms, the prototypical class of exponentially difficult to optimize functions, may be expressed in this form.
Therefore, we do not expect to find an optimizer which can efficiently deal with all members of this class, and so cannot globally optimize any acquisition function depending on the predictive variance.
Incidentally, the posterior mean is an additive function, and can itself thus be efficiently optimized.

Because of this difficulty, it has long been proposed to discard the bilateral terms $\sigma^2_{i,j}$ for $i\neq j$.
\citet{kandasamy2015high} themselves proposed an additive approximation to the Lower Confidence Bound (aLCB) method: 
\begin{equation}
    \alpha_{aLBC}^\beta(\x) = 
    \sum_{p=1}^P \mu_p(x_p) - \beta \sigma_p(x_p) \,,
\end{equation}
that is, aLCB is given by the sum of the LCB objectives associated marginally with each subfunction.
\cite{kandasamy2015high} have shown that asymptotically, there is not much lost by their approximation, which includes dropping the bilateral uncertainty, iin terms of a regret analysis.

\citet{mutny2018efficient} show that a Fourier Feature approximation to the GP allows for a Thompson Sampling \citep{thompson1933likelihood} approach which respects bilateral uncertainty.
One purpose of our paper is to show that no such Fourier Feature approximation is necessary. 
As we make precise in Section \ref{sec:indep}, though $f_i$ and $f_j$ are not independent \textit{a posteriori}, they are conditionally independent given the difference between the observed black-box outputs and the value of $f_j$ at the input locations.
This suggests a sequential method applicable to any additive kernel which allows for  efficient Thompson Sampling, which is of methodological interest.

It is also of substantive interest.
Most acquisition functions, such as LCB or max-value entropy \citep{wang2017max}, are not possible to reliably globally optimize when accounting for bilateral uncertainty.
This methodology thus provides a unique opportunity to conduct an empirical study of the extent to which the theoretical suggestions that bilateral uncertainty can be ignored are borne out on real problems with small budgets.
By comparing the result of the additive and exact methods, we thus measure the advantage, if any, of incorporating bilateral uncertainty.

In our experiments, we will find that there is not a consistent and practically significant advantage in incorporating bilateral uncertainty in Thompson Sampling for additive BO.
Though this finding must be subject to caveats including the particular functions that we used for benchmarking, our particular implementation of BO, and the focus on Thompson Sampling, we think it still suggests broader takeaways for the additive BO methodology community, as it suggests that computational and research effort expended to account for bilateral uncertainty may not be of the highest priority.

The rest of our paper is organized as follows. 
In Section \ref{sec:sota} we further discuss related work on additive Bayesian optimization.
We present our methodological contributions in Section \ref{sec:methods}.
Next, we present our substantive study results in Section \ref{sec:results} before providing discussion and conclusions in Section \ref{sec:conclusion}.

\begin{figure}
    \centering
    \includegraphics[width=0.98\linewidth]{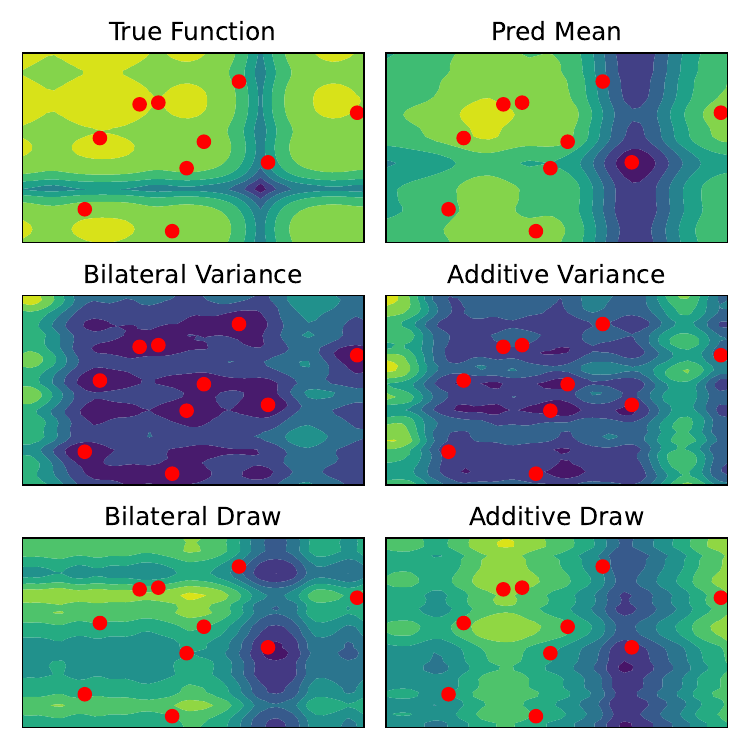}
    \caption{\textbf{Illustration of Additive Thompson Sampling.}
    Top left panel gives true response surface; top right gives posterior mean of an additive GP.
    The middle row gives the predictive variance of an additive GP (left) and its approximation without bilateral terms (right).
    The bottom row gives a predictive draw from each.
    }
    \label{fig:dim2}
\end{figure}

\section{The State of the Art in Additive Bayesian Optimization}
\label{sec:sota}

We begin by establishing our notation and problem setting in Section \ref{sec:setting} before providing an overview of recent relevant literature in Section \ref{sec:literature}.
Finally, we provide a summary of previous discussion of bilateral uncertainty in additive BO in Section \ref{sec:bibg}.

\subsection{Problem Setting and Notation}
\label{sec:setting}

This article is interested in the problem of Black Box Optimization, that is, we posit a function $f:[0,1]^P \to\mathbb{R}$ of which we would like to find the global minimum.
We assume that for any $\x\in[0,1]^P$, we can evaluate a noise-corrupted version $y = f(\x)+\epsilon_i$ at that point and that evaluations are costly enough that we can only afford a few.
The \textit{Additive Assumption} consists of imposing on $f$ the structure $f(\x) = \sum_{m=1}^{M} f_i(\P_m\x)$, which breaks the function up as a sum of lower dimensional functions, each acting on a subset $\I_m\subset\{1,\ldots,P\}$ of the variables
We denote the dimension of each subset by $\sdim_m$, that is, $\sdim_m = |\I_m|$.
The matrix $\P_m\in\mathbb{R}^{\sdim_m\times P}$ picks out the coordinates of $\x$ in $\I_m$ such that $\P_m\x\in\mathbb{R}^{\sdim_m}$.
We assume that $\epsilon\overset{iid}{\sim}N(0,\tau^2)$; this might represent either the randomness of a stochastic black-box, or it might represent the discrepancy between a true deterministic black-box and the additive assumption.

There are two equivalent ways of conceptualizing Gaussian processes within the additive framework \citep{durrande2011additive,duvenaud2011additive}. 
First, we can define a single GP on the black-box $f\sim GP(\mu,\k)$, but use a compound kernel which consists of a sum of kernels over the individual variable blocks: 
$\k(\x^1,\x^2) = \sum_{m=1}^M \k_m(\P_m\x^1, \P_m\x^2)$.
Alternatively, we can assign to each subfunction $f_i$ a Gaussian process $f_i \sim GP(\mu_i, \k_i)$ using a standard kernel $k_i$, and subsequently define $f(\x) = \sum_{i=1}^M f_m(\P_m \x)$.
Either of these definitions implies the same distribution on $f$.
We will find this second view to be more fruitful for this article.
For $\A\in\mathbb{R}^{N_a\times \sdim_i}$, $\B\in\mathbb{R}^{N_b\times \sdim_i}$, we will denote by $\K_i(\A,\B)\in\mathbb{R}^{N_a\times N_b}$ the matrix with $j,l$ entry given by the kernel $k_i$ applied to the $j^{\textrm{th}}$ row of $\A$ and $l^{\textrm{th}}$ row of $\B$.

Given a surrogate model, it remains to specify a rule for using it to choose points to evaluate sequentially.
The rule which is the focus of this article is Thompson Sampling \cite{thompson1933likelihood}.
This is a stochastic rule for sequential learning which specifies that we should choose a point randomly with probability proportional to its being the optimal one given the current state of knowledge.
In the context of GPs, this is typically done by specifying some finite set of points $\mathcal{X}\in[0,1]^P$, sampling the posterior predictive at those points, and then selecting the minimum value achieved on $\mathcal{X}$.
For a two dimensional additive Gaussian Process, we have the following posterior predictive given $(\X,\y)$, where we stack the elements of $\mathcal{X}$ into rows of a matrix $\X'$:
\begin{align*}
    & f(\X') = f_1(\P_1\X')+f_2(\P_2\X') \sim N(\boldsymbol{\mu}, \boldsymbol{\Sigma})\,, \\
    & \boldsymbol{\mu} = \bar{\K}(\X',\X)\big(\bar{\K}(\X,\X)+\tau^2\mathbf{I}\big)^{-1}\y \,,
    \\ &
    \boldsymbol{\Sigma} = \bar{\K}(\X',\X') - \bar{\K}(\X',\X)\big(\bar{\K}(\X,\X)+\tau^2\mathbf{I}\big)^{-1}\bar{\K}(\X,\X') \,,
    \\ & 
    \textrm{where } \bar{\K}(\A,\B) = \K_1(\P_1\A,\P_1\B)+\K_2(\P_2\A,\P_2\B) \,.
\end{align*}
The second term of $\boldsymbol\Sigma$ (after the minus) breaks up into:
\begin{align}
    \sum_{i,j=1}^2 K_i(\P_i\X',\P_i\X)(\bar{\K}(\X,\X)+\tau^2\mathbf{I})^{-1}K_j(\P_j\X,\P_j\X')\,.
\label{eq:truesigma}
\end{align}
The additive approximation refers to the distribution which simply drops the terms for $i\neq j$ (i.e. the bilateral terms):
\begin{align}
    \sum_{i=1}^2 K_i(\P_i\X',\P_i\X)(\bar{\K}(\X,\X)+\tau^2\mathbf{I})^{-1}K_i(\P_i\X,\P_i\X')\,.
    \label{eq:addsigma}
\end{align}
Figure \ref{fig:dim2} shows an example using a GP fit on a sum of Ackley functions for each dimension (see Appendix \ref{sec:funcs} for precise definition). 
The mid and bottom left panes show the posterior variance surface and a draw from the exact posterior, respectively, which respect the bilateral uncertainty (i.e. using Equation \ref{eq:truesigma}).
The mid and bottom right panes show the additive variant (i.e. using Equation \ref{eq:addsigma}).
Qualitatively, they look very similar, though we notice that the variance including bilateral uncertainty does a better job of vanishing near design points.

\subsection{Recent Work in Additive GP Modeling}
\label{sec:literature}
\citet{durrande2011additive} and \citet{duvenaud2011additive} were early proposers of additive kernels.
Though already, these authors studied subkernels involving more than one variable both of these works study them in the regression context rather than the sequential design/active learning context.
This was the contribution of \citet{kandasamy2015high}, who proposed the use of additive GPs for BO.
Furthermore, \citet{kandasamy2015high} also proposed the use of the additive LCB method which uses as a surrogate $\sum_{m=1}^M \mu_m(\P_m\x)-\beta\sigma_m(\P_m\x)$, i.e. which throws away the bilateral uncertainty.

So far, we have been assuming an additive structure, that is to say, assuming knowledge of the correct $M$ and $\P_m$ to use.
While sometimes known in practice (e.g. the WLAN problem studied by \citet{bardou2023relaxing}), under black-box assumptions this has to be inferred.
\citet{gardner2017discovering} suggested Bayesian learning of the decomposition via posterior simulation, and \citet{han2021high} proposed a tree-based method for decomposition learning.
On the other hand, \cite{ziomek2023random} advocate to just use random decompositions, and justify this theoretically in the context of adversarially chosen decompositions.

Recent years have seen more advanced methods published within the umbrella of additive BO that allow for more complex structure.
\citet{wang2017max} and \citet{wang2017batched} studied alternative acquisition functions from information theory, finding them to be profitable in the additive BO context.
\citet{rolland2018high,hoang2018decentralized} proposed message-passing algorithms and \citet{bardou2023relaxing} proposed a convex-optimization-based algorithm for acquisition when the subfunction indices overlap, which is a more complex problem as the optimization subproblems are coupled and cannot simply be solved independently. 
\citet{lu2022additive} show that in the context of overlapping additive functions, orthogonalizing them against one another can be beneficial.

\subsection{Previous Discussion of Bilateral Uncertainty}
\label{sec:bibg}

Ignoring posterior correlation between the subfunctions has been standard practice since \citet{kandasamy2015high} proposed it in the very paper establishing additive BO.
\cite{wang2017batched} generates individual coordinates for each group then combines them by random sampling or so as to maximize the sum of the individual acquisitions.
An exception is \cite{mutny2018efficient}, who show how to respect them in the context of Thompson Sampling with the Fourier Features \citep{rahimi2007random} approximation to Gaussian processes.
Beyond ignoring these correlations, the additive decomposition of the LCB combines variance incorrectly by adding the standard deviations rather than the variance.
\citet{bardou2023relaxing} suggest using a different measure of spread to improve this approximation.
\citet{wang2017max} propose an algorithm for max-value entropy search which combines marginal optima within each subfunction, which discards bilateral uncertainty.

\section{Thompson Sampling on Additive Domains}
\label{sec:methods}

In this section we discuss three strategies for producing posterior samples from the subfunctions $f_m$ evaluated at arbitrary points in their domain.
First, in Section \ref{sec:naive}, we discuss a naive approach which constructs the joint posterior between all $M$ subfunctions.
Next, we discuss the independent approximation in Section \ref{sec:marg} where sampling is conducted from each marginal posterior.
In Section \ref{sec:indep} we examine the joint distribution of an augmented process finding useful conditional independence, and use this to propose an efficient autoregressive method for joint posterior draws in Section \ref{sec:joint}.

\subsection{The Joint Posterior}
\label{sec:naive}

We begin by defining some notation for key quantities parameterizing the multivariate normal distributions of interest.
We assume that we have some candidate points 
$\mathcal{Z} = \Z_1 \times \Z_2 \times \ldots \times \Z_M
\in
\mathbb{R}^{B_1\times\sdim_1}
\times
\mathbb{R}^{B_2\times\sdim_2}
\times
\ldots
\times
\mathbb{R}^{B_M\times\sdim_M}
$
.
Note that we can have a separate number of points $B_m$ associated with each subdomain $m$; we may wish to use more for those of higher dimension.
Now define $\Ktall\in\mathbb{R}^{N\times \sum_{m=1}^M B_m}$ as the matrix containing the covariances between the predictive and training locations for each subdomain; that is:
\begin{align*}
    \Ktall = 
    \begin{bmatrix}
        \K_1(\P_1\X,\Z_1) &
        \ldots &
        \K_M(\P_M\X,\Z_M) &
    \end{bmatrix}^\top \,.
\end{align*}
Now, let $\Knew\in\mathbb{R}^{\sum_{m=1}^M B_m \times\sum_{m=1}^M B_m} $ be the block-diagonal matrix giving the prior covariance for all the subdomains put together:
\begin{align*}
    \Knew = 
    \begin{bmatrix}
        \K_1(\Z_1, \Z_1) & \zero & \ldots & \zero\\
        \zero & \K_2(\Z_2, \Z_2) & \ldots & \zero\\
        \vdots & \vdots & \ddots & \vdots \\
        \zero & \zero & \ldots & \K_M(\Z_M, \Z_M) \\
    \end{bmatrix} \,.
\end{align*}

If we could sample each $f_m$, we could execute Thompson Sampling via a grid search for sufficiently small $\sdim_m$.
To this end, we can develop a joint posterior on 
$[f_1(\Z_1), f_2(\Z_2), \ldots, f_M(\Z_M)]\sim N(\boldsymbol\mu^\mathcal{Z}, \boldsymbol\Sigma^\mathcal{Z})$:
\begin{align*}
& \boldsymbol\mu^{\mathcal{Z}} = 
\Ktall
^\top
\Bigg(
\sum_{j=1}^J 
\K_j(\P_j\X,\P_j\X)
+ \tau^2 \mathbf{I}
\Bigg)^{-1}
\y \,,\\
& \boldsymbol\Sigma^{\mathcal{Z}} = 
\Knew - 
\Ktall
^\top
\Bigg(
\sum_{j=1}^J 
\K_j(\P_j\X,\P_j\X)
+ \tau^2 \mathbf{I}
\Bigg)^{-1}
\Ktall \,.
\end{align*}
$\boldsymbol\Sigma^{\mathcal{Z}}$ is of dimension $\sum_{m=1}^M B_m \leq M B^*$ where $B^* = \max_{m\in\{1,\ldots,M\}} B_m$.
The most straightforward technique for sampling from a multivariate normal involves computing a Cholesky decomposition which is of complexity cubic in matrix dimension, making this strategy of complexity $M^3B^{*3}$. 
Clearly, this is a prohibitive approach for even moderate $M$.
One of our tasks in this article is to tame this computational complexity.
We first consider a linear algebraic approach.
Notice that $\boldsymbol\Sigma^{\mathcal{Z}}$ is a rank $N$ update of a $MB^*\times MB^*$ block-diagonal matrix.
This suggests the use of a low-rank Cholesky update \citep{seeger2004low}, since the Cholesky factor for the block-diagonal matrix can be computed in $MB^{*3}$ time, and each of the $M$ Cholesky subfactors can be computed in parallel.
However, the update step is still quite costly, on the order of $N^{2}M^2$, and the updates must be performed sequentially, making this approach ill-suited for modern computing environments.
We found that this was actually slower than naively calculating the Cholesky decomp on a modern GPU (we used a 1080 Ti).

\subsection{Marginal Thompson Sampling}
\label{sec:marg}

One way to avoid this large computation would be to take a page from most existing additive acquisition strategies and ignore the bilateral correlations by sampling each subfunction independently.
The marginal posterior on $f_m$ give $\y$, which is well known (see e.g. \citep{wang2017max}), is given  by
$f_i(\Z_i)|\y \sim 
N(\bar\mu_i,\bar\Sigma_i)
$
where
\begin{align*}
    & \bar\mu_i = 
    \K_i(\Z_i,\P_i\X)
    \tilde{\K}^{-1}
    \y \,,
    \\
    & \bar\Sigma_i = 
    \K_i(\Z_i,\Z_i) - 
    \K_i(\Z_i,\P_i\X) 
    \tilde{\K}^{-1}
    \K_i(\P_i\X_i,\Z_i)  \,,
    \\ &
    \tilde{\K} = \Bigg(
    \sum_{j=1}^J 
    \K_j(\P_j\X,\P_j\X)
    + \tau^2 \mathbf{I}
    \Bigg) \,.
\end{align*}
This allows for sampling each subfunction independently of one another (see Algorithm \ref{alg:indep}).
This can be done in parallel, and the cost is dominated by the Cholesky decomposition, such that the total cost in serial is on the order of $\sum_{m=1}^M B_m^3$ and with ideal parallelism we would observe on the order of $B^{*3}$ computing time required.
However, it comes at the cost of ignoring the bilateral uncertainty.

\begin{algorithm}[tb]
   \caption{Marginal Thompson Sampling}
   \label{alg:marg}
\begin{algorithmic}
    \REQUIRE{Training Data $(\X,\y)$, Sampling locations 
    $\Z_1\times\Z_2,\ldots,\Z_M \in \mathbb{R}^{\sdim_1,P}\times\mathbb{R}^{\sdim_2,P}\times\ldots\times\mathbb{R}^{\sdim_M,P}$}.
   \STATE  $\tilde{\K} \gets  \sum_{m=1}^M\K_m(\P_m \X,\P_m \X) + \tau^2 \mathbf{I}$
   \FOR{$m=1$ {\bfseries to} $M$}
   \STATE $\bar{\boldsymbol{\mu}}_i \gets  \K_i(\Z^i,\P_m\X) \tilde{\K}^{-1} \y $
   \STATE $\bar{\boldsymbol{\Sigma}}_i \gets   \K_i(\Z^i,\Z^i) -  \K_i(\Z^i,\P_i\X)  \tilde{\K}^{-1} \K_i(\P_i\X^i,\Z^i)$
   \STATE $\boldsymbol\xi\sim N(0,\I)$
   \STATE $\bar{\mathbf{L}}_m \gets \textrm{cholesky}(\bar{\boldsymbol\Sigma}_m)$
   \STATE $\mathbf{f}^z_m \gets \bar{\mathbf{L}}_m\boldsymbol{\xi} + \bar{\boldsymbol{\mu}}_m$
   \ENDFOR
   \RETURN Samples $\mathbf{f}_1^z, \mathbf{f}_2^z, \ldots, \mathbf{f}_M^z$ at $\Z_1,\Z_2,\ldots,\Z_m$.
\end{algorithmic}
\label{alg:indep}
\end{algorithm}

\subsection{The Residual Conditionals}
\label{sec:indep}

Recall that though the functions $\{f_1,f_2,\ldots,f_M\}$ are independent \textit{a priori}, we observe only their sum at any given point and consequently they are dependent \textit{a posteriori}.
However, it turns out that we do have conditional independence of any two processes if we condition not on the observed data $\y$, but on the residuals of the data given one of the two processes at the design points; that is, $\y-f_j(\P_j\X)$.
\begin{proposition}
    \label{prop:indep}
    For any $\Z^i\in\mathbb{R}^{B_i\times\sdim_i}$ 
    and $\Z_j\in\mathbb{R}^{B_j\times\sdim_j}$, we have that
    $f_i(\Z_i) \indep f_j(\Z_j) | \y - f_j(\P_j \X)$.
\end{proposition}
This simple fact may be established straightforwardly using typical multivariate Normal computations.
However, its useful consequences for additive BO do not appear to previously have been described.

\begin{figure}[b]
    \centering
    \begin{tabular}{|c|}
        \hline
        Samples conditional on...\\
        \includegraphics[width=0.98\linewidth]{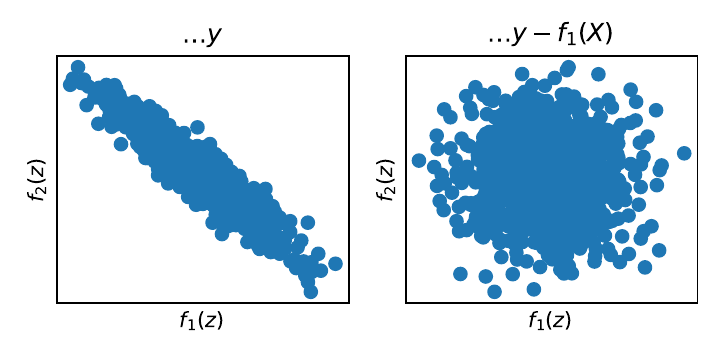}\\
        \hline
    \end{tabular}
    
    \caption{\textbf{Example joint distribution in 2D.} Left shows posterior conditional, right shows conditional on residuals.}
    \label{fig:indep}
\end{figure}

Before describing them, however, we illustrate the two dimensional case by considering the conditional distribution of $f_2(\Z_2),f_1(\Z_1)|\y - f_1(\P_1\X)$.
Straightforward arithmetic gives us that :
\begin{align*}
    &
    \begin{bmatrix}
        f_1(\Z_1) \\f_2(\Z_2)
    \end{bmatrix}
    | \y-f_1(\P_1\X)
    \sim
    N (
    \hat{\boldsymbol\mu},\hat{\boldsymbol\Sigma}); \hspace{2em} \textrm{with}
    \\ &
    \hat{\boldsymbol\mu} = 
    \begin{bmatrix}
        \mathbf{0}_{B_1\times N} \\
        K_2(\Z_2,\X)
    \end{bmatrix}
    \Big(
    K_2(\P_2\X,\P_2\X) + \tau^2 \mathbf{I}
    \Big)
    (\y-f_1(\X)) 
    \\ & 
    \textrm{and } \hat{\boldsymbol\Sigma} = 
    \begin{bmatrix}
        K_1(\Z_1,\Z_1) & \mathbf{0}_{B_1\times B_2} \\
        \mathbf{0}_{B_2\times B_1} & 
        K_2(\Z_2,\Z_2) - \K^*
    \end{bmatrix}\,,
\end{align*}
where $\K^* $ is given by:
\begin{equation*}
    \K_2(\Z_2,\P_2\X) 
    \Big(
    \K_2(\P_2\X,\P_2\X) + \tau^2 \mathbf{I}
    \Big)^{-1}
    \K_2(\P_2\X,\Z_2)  \,.
\end{equation*}
We notice that the conditional covariance is block-diagonal, indicating independence.
Figure \ref{fig:indep} contrasts this conditional independence with the strong dependence found in the conditional distribution given $\y$, i.e. the posterior.

Moving to the general case, an immediate consequence of Proposition \ref{prop:indep} is that 
\begin{equation*}
     f_i(\Z_i) \indep [f_{1}(\Z_1), \ldots, f_{i-1}(\Z_{i-1})] | \y - \sum_{j=1}^{i-1} f_j(\P_j\X)   \,,
\end{equation*}
i.e., if we have some subfunction $i$ and a collection of different subfunctions $\{1, \ldots, i-1\}$, then if we condition on the residuals considering only those second subfunctions, we have independence.
Furthermore, the marginal distribution of $f_i(\Z_i)$ conditional on the residuals is given by 
$f_i(\Z_i)|\y-\sum_{j=1}^{i-1} f_j(\P_j\X) \sim 
N(\tilde{\boldsymbol\mu}_i,\tilde{\boldsymbol\Sigma}_i)
$
where
\begin{align}
    & \tilde{\boldsymbol\mu}_i = 
    \K_i(\Z_i,\P_i\X)
    \tilde{\K}_i^{-1}
    (\y - \sum_{j=1}^{i-1} f_j(\P_j\X))
    \label{eq:mut} \,,
    \\
    & \tilde{\boldsymbol\Sigma}_i = 
    \K_i(\Z_i,\Z_i) - 
    \K_i(\Z_i,\P_i\X) 
    \tilde{\K}_i^{-1}
    \K_i(\P_i\X,\Z_i)  \,,
    \label{eq:sigt}
\end{align}
and
\begin{align}
    & \tilde{\K}_i =
    \sum_{j=1}^{i-1} 
    \K_j(\P_j \X,\P_j \X)
    + \tau^2 \mathbf{I} \,.
\end{align}
Here, we used the convention that a sum with $i-1<j$ is empty, such that $\tilde{\K}_1=\tau^2\mathbf{I}$ and $\tilde{\boldsymbol\mu}_1 = 
\K_1(\Z_1,\P_1\X)
\tilde{\K}_1^{-1}
\y $.

\subsection{Efficient Joint Thompson Sampling}
\label{sec:joint}

This conditional independence of the previous section suggests a sequential strategy for jointly sampling $[f_1(\Z_1), \ldots, f_M(\Z_M)]$ by sampling the augmented vector $[f_1(\Z_1), f_1(\P_1\X), \ldots, f_M(\Z_M), f_M(\P_M\X)]$ autoregressively, conditioning on $\y - \sum_{j=1}^{i-1} f_{j}(\X)$ at each step $i$.
That is, to conduct the sampling scheme:
\begin{align*}
    & [f_1(\Z_1),f_1(\P_1\X)] \sim N(\tilde{\boldsymbol\mu}_1, \tilde{\boldsymbol{\Sigma}_1}) \\
    & [f_2(\Z_2),f_2(\P_2\X)] | \y-f_1(\P_1\X) \sim N(\tilde{\boldsymbol\mu}_2, \tilde{\boldsymbol{\Sigma}}_2) \\
    & \hspace{10em}\vdots \\
    & [f_M(\Z_M),f_M(\P_M\X)] | \y-\sum_{m=1}^{M-1}f_m(\P_m\X) \sim N(\tilde{\boldsymbol\mu}_M, \tilde{\boldsymbol{\Sigma}}_M)  \,,
\end{align*}
where for $i>1$, $\tilde{\boldsymbol{\mu}}_i$ depends on $\y-\sum_{j=1}^{i-1}$, and $\tilde{\boldsymbol{\mu}},\tilde{\boldsymbol{\Sigma}}$ are as in Equations \ref{eq:mut} and \ref{eq:sigt}.
Subsequently, we simply discard $[f_1(\P_1\X), \ldots, f_M(\P_M\X)]$, the samples at the design points.
We provide pseudocode delineating further details in Algorithm $\ref{alg:main}$.

\begin{algorithm}[tb]
   \caption{Efficient Additive Thompson Sampling}
   \label{alg:main}
\begin{algorithmic}
    \REQUIRE{Training Data $\X,\y$, Sampling locations 
    $\Z_1\times\Z_2,\ldots,\Z_M \in \mathbb{R}^{\sdim_1,P}\times\mathbb{R}^{\sdim_2,P}\times\ldots\times\mathbb{R}^{\sdim_M,P}$}.
   \FOR{$m=1$ {\bfseries to} $M$}
   \STATE $\tilde{\K}_m \gets \sum_{j=1}^{m-1} \K_j(\P_j\X,\P_j\X) + \tau^2$
   \STATE $\mathbf{B}_m \gets \tilde{\K_m}^{-1} \K_m(\P_m\X,[\P_m\X,\Z_m])$
   \STATE $\tilde{\boldsymbol{\mu}}_m \gets \mathbf{B}^{m\top} (\y - \sum_{j=1}^{m-1} \mathbf{f}_j^x)$
   \STATE $\tilde{\Sigma}_m \gets \K_m([\P_m\X,\Z_m],[\P_m\X,\Z_m])  - \K_m([\P_m\X,\Z_m],\P_m\X) \mathbf{B}_m 
   $
   \STATE $\boldsymbol\xi\sim N(0,\I)$
   \STATE $\tilde{\mathbf{L}}_m \gets \textrm{cholesky}(\tilde{\Sigma}_m)$
   \STATE $[\mathbf{f}^x_m,\mathbf{f}^z_m] \gets \tilde{\mathbf{L}}_m\boldsymbol{\xi} + \tilde{\boldsymbol{\mu}}_m$
   \COMMENT{First $N$ elements in $\mathbf{f}^x_m$.}
   \ENDFOR
   \RETURN Samples $\mathbf{f}_1^z, \mathbf{f}_2^z, \ldots, \mathbf{f}_M^z$ at $\Z_1,\Z_2,\ldots,\Z_m$.
\end{algorithmic}
\end{algorithm}

As before, the most computationally intensive part is the Cholesky decomposition, which requires in total on the order of $M(N+B^*)^3$ operations. 
Unlike the marginal sampling of Section \ref{sec:marg}, this sampling is not embarrassingly parallel, as in order to generate the samples for $f_m$ we need to know the samples of $\{f_1,\ldots,f_{m-1}\}$ at $\X$. 
However, this serial dependence only affects the mean part of the sampling, not the covariance.
Therefore, an efficient implementation will first form $\tilde{\mathbf{L}}_i\boldsymbol{\xi}_i$ where $\tilde{\boldsymbol{\Sigma}}=\tilde{\mathbf{L}}\tilde{\mathbf{L}}^\top$ in parallel for each $i$ before adding the mean term serially.
Since the Cholesky decomposition dominates the complexity of the sampling, we can observe $(N+B^{*})^3$ complexity with ideal parallelism.
Thus, so long as we are in the small sample size case, it will be of similar computational expense as the independent sampler.
In any event, it requires on the order of $M^2$ times less operations than the naive joint sampling approach of Section \ref{sec:naive}.
Furthermore, batch sampling can be done also in parallel and without any additional decompositions;
see Appendix \ref{sec:numpy} for Numpy \citep{harris2020array} code implementing the efficiency considerations just discussed.

\begin{figure*}
    \centering
    \begin{tabular}{|c|c|}
        \hline
        \textbf{Batch Size 1} & \textbf{Batch Size 10} \\
        \hline
        \includegraphics[width=0.47\linewidth,trim={0.50em 1.25em 1.05em 0.75em},clip]{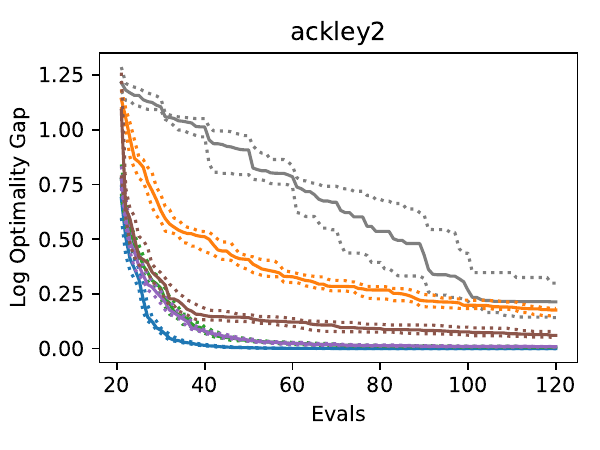}
        &
        \includegraphics[width=0.47\linewidth,trim={0.50em 1.25em 1.05em 0.75em},clip]{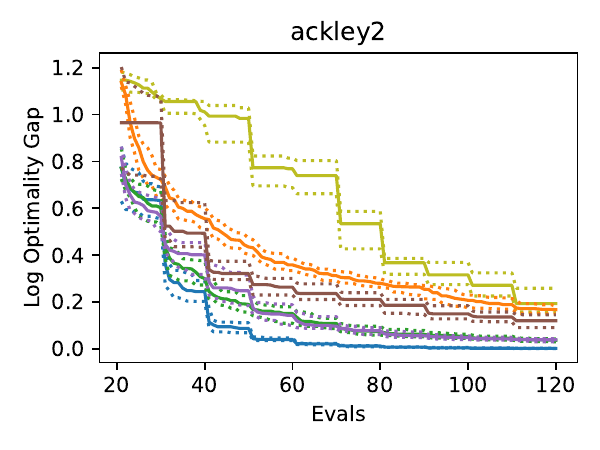}
        \\ 
        \hline
        \includegraphics[width=0.47\linewidth,trim={0.50em 1.25em 1.05em 0.75em},clip]{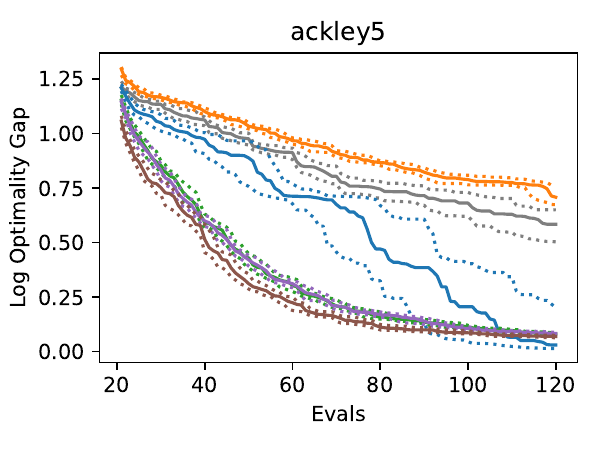}
        &
        \includegraphics[width=0.47\linewidth,trim={0.50em 1.25em 1.05em 0.75em},clip]{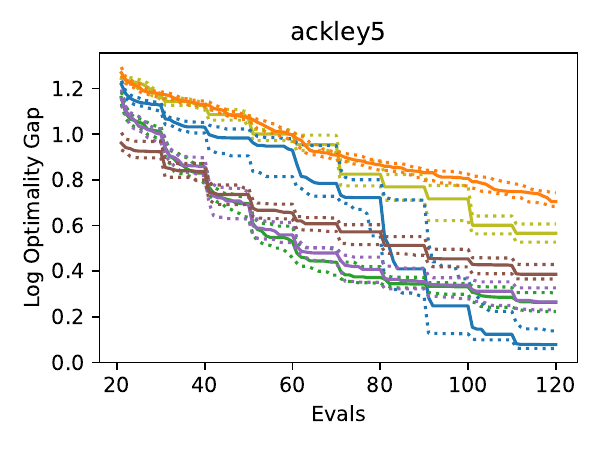}
        \\ 
        \hline
        \includegraphics[width=0.47\linewidth,trim={0.50em 1.25em 1.05em 0.75em},clip]{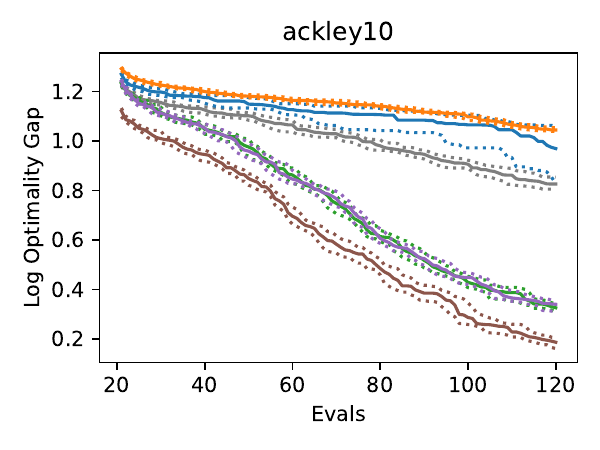}
        &
        \includegraphics[width=0.47\linewidth,trim={0.50em 1.25em 1.05em 0.75em},clip]{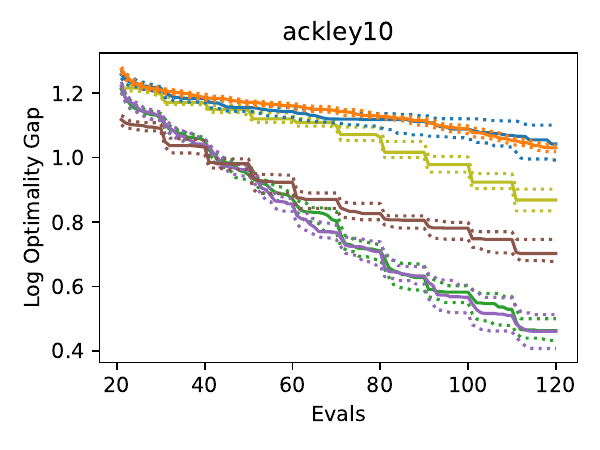}
        \\ 
        \hline
        \multicolumn{2}{|c|}{\includegraphics[width=0.99\linewidth]{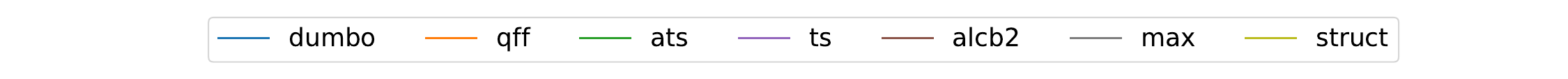}} \\
        \hline
    \end{tabular}
    
    \caption{\textbf{Results for Random Additive Ackley functions}. Solid line gives median performance and dotted lines give an asymptotic 95\% confidence interval thereon. These results are aggregated over 100 replicates.}
    \label{fig:ackley}
\end{figure*}

\begin{figure*}
    \centering
    \begin{tabular}{|c|c|}
        \hline
        \textbf{Batch Size 1} & \textbf{Batch Size 10} \\
        \hline
        \includegraphics[width=0.47\linewidth,trim={0.50em 1.25em 1.05em 0.75em},clip]{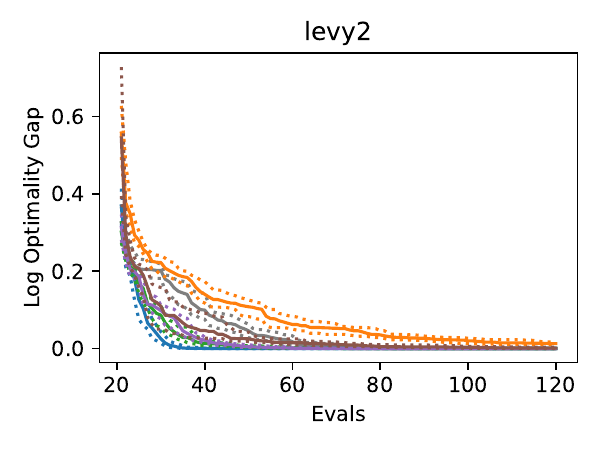}
        &
        \includegraphics[width=0.47\linewidth,trim={0.50em 1.25em 1.05em 0.75em},clip]{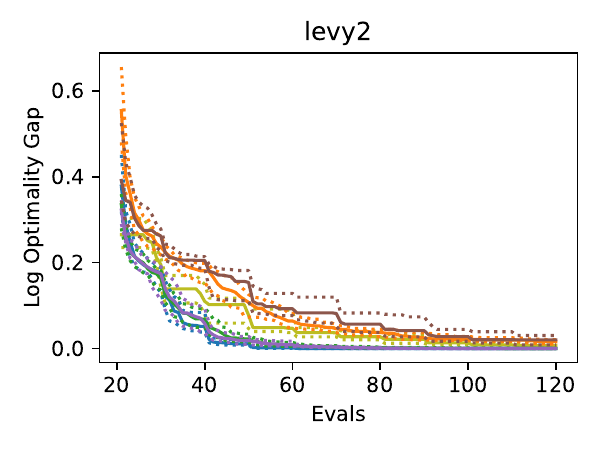}
        \\ 
        \hline
        \includegraphics[width=0.47\linewidth,trim={0.50em 1.25em 1.05em 0.75em},clip]{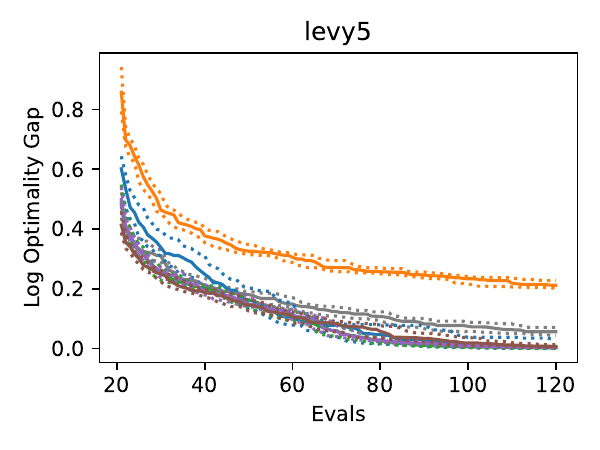}
        &
        \includegraphics[width=0.47\linewidth,trim={0.50em 1.25em 1.05em 0.75em},clip]{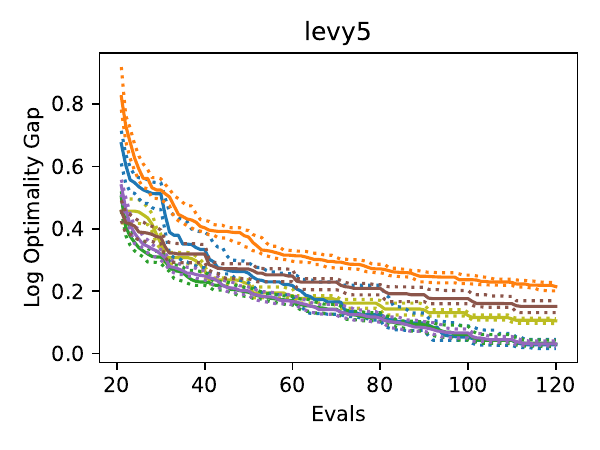}
        \\ 
        \hline
        \includegraphics[width=0.47\linewidth,trim={0.50em 1.25em 1.05em 0.75em},clip]{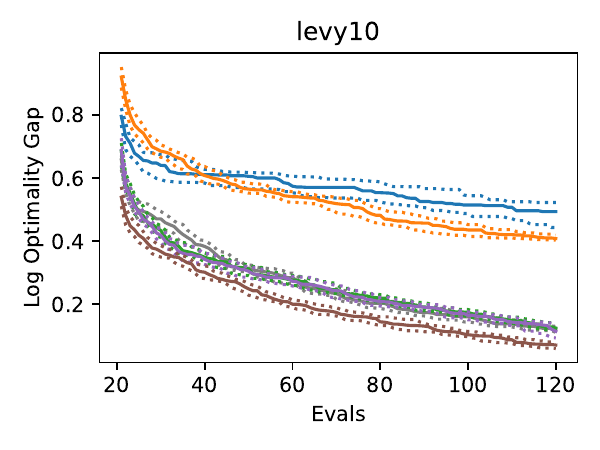}
        &
        \includegraphics[width=0.47\linewidth,trim={0.50em 1.25em 1.05em 0.75em},clip]{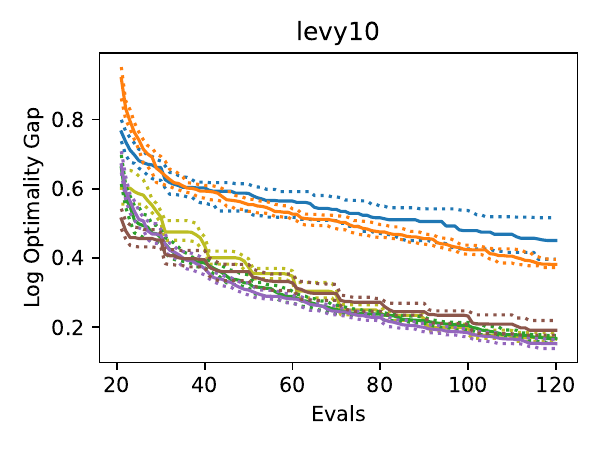}
        \\ 
        \hline
        \multicolumn{2}{|c|}{\includegraphics[width=0.99\linewidth]{images/legend.pdf}} \\
        \hline
    \end{tabular}
    
    \caption{\textbf{Results for Random Additive levy functions}. Solid line gives median performance and dotted lines give an asymptotic 95\% confidence interval thereon. These results are aggregated over 100 replicates.}
    \label{fig:levy}
\end{figure*}

\begin{figure*}
    \centering
    \begin{tabular}{|c|c|}
        \hline
        \textbf{Batch Size 1} & \textbf{Batch Size 10} \\
        \hline
        \includegraphics[width=0.47\linewidth,trim={0.50em 1.25em 1.05em 0.75em},clip]{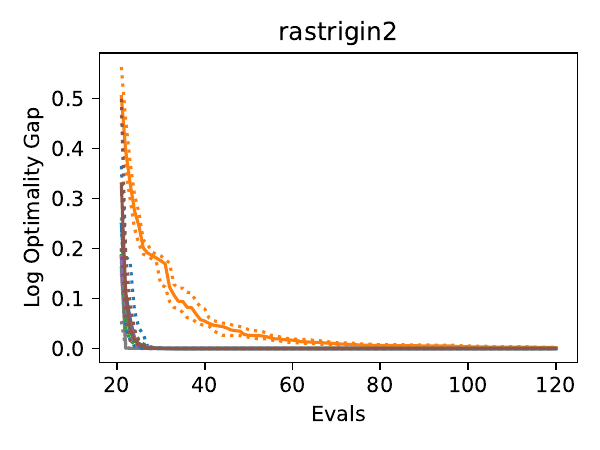}
        &
        \includegraphics[width=0.47\linewidth,trim={0.50em 1.25em 1.05em 0.75em},clip]{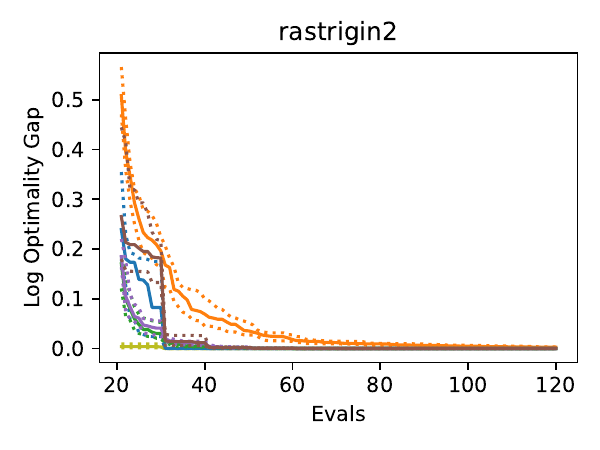}
        \\ 
        \hline
        \includegraphics[width=0.47\linewidth,trim={0.50em 1.25em 1.05em 0.75em},clip]{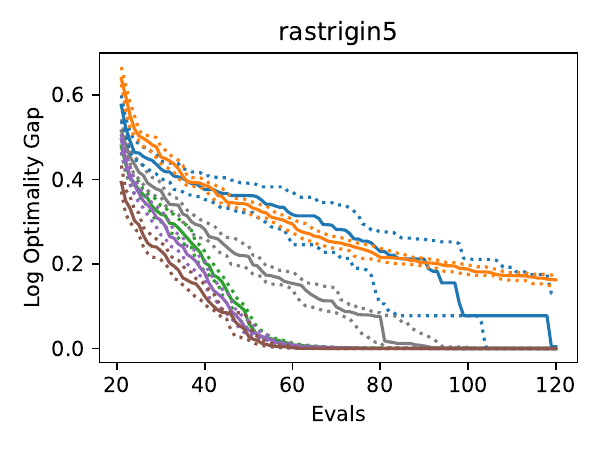}
        &
        \includegraphics[width=0.47\linewidth,trim={0.50em 1.25em 1.05em 0.75em},clip]{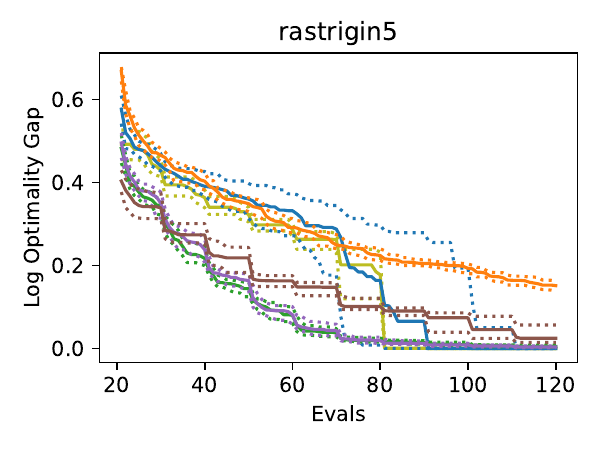}
        \\ 
        \hline
        \includegraphics[width=0.47\linewidth,trim={0.50em 1.25em 1.05em 0.75em},clip]{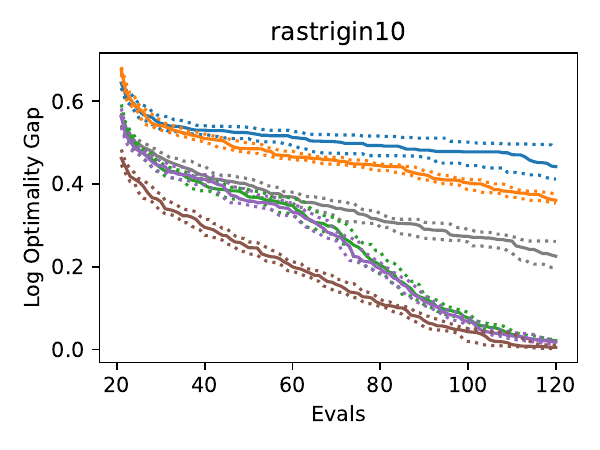}
        &
        \includegraphics[width=0.47\linewidth,trim={0.50em 1.25em 1.05em 0.75em},clip]{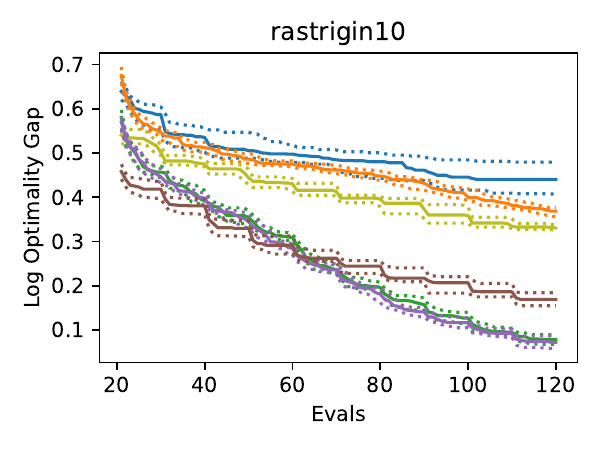}
        \\ 
        \hline
        \multicolumn{2}{|c|}{\includegraphics[width=0.99\linewidth]{images/legend.pdf}} \\
        \hline
    \end{tabular}
    
    \caption{\textbf{Results for Random Additive rastrigin functions}. Solid line gives median performance and dotted lines give an asymptotic 95\% confidence interval thereon. These results are aggregated over 100 replicates.}
    \label{fig:rastrigin}
\end{figure*}

\begin{figure*}
    \centering
    \begin{tabular}{|c|c|}
        \hline
        \textbf{Batch Size 1} & \textbf{Batch Size 10} \\
        \hline
        \includegraphics[width=0.47\linewidth,trim={0.50em 1.25em 1.05em 0.755em},clip]{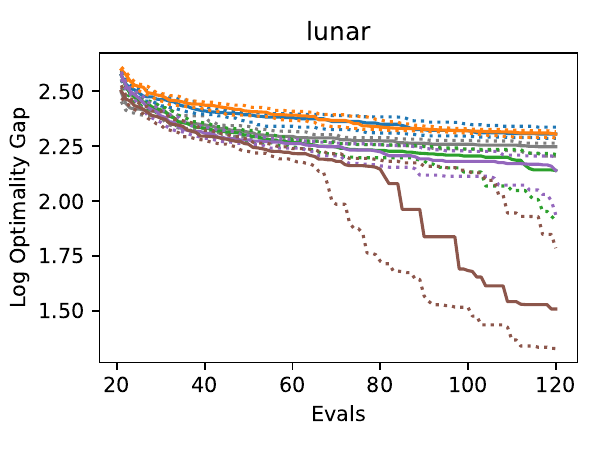}
        &
        \includegraphics[width=0.47\linewidth,trim={0.50em 1.25em 1.05em 0.755em},clip]{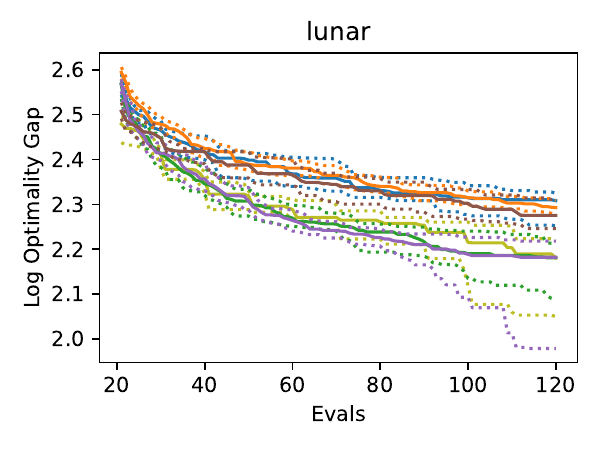}
        \\ 
        \hline
        \includegraphics[width=0.47\linewidth,trim={0.50em 1.25em 1.05em 0.755em},clip]{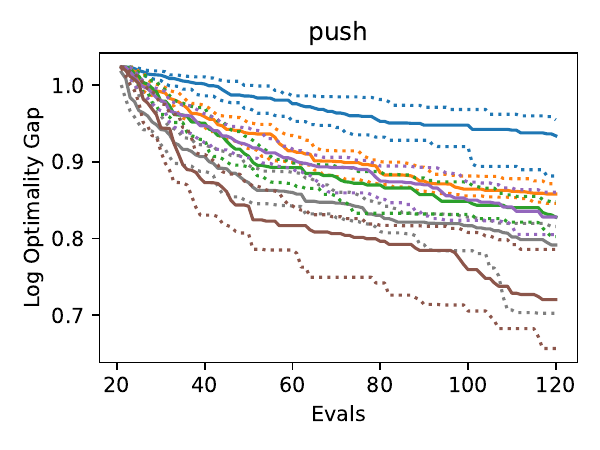}
        &
        \includegraphics[width=0.47\linewidth,trim={0.50em 1.25em 1.05em 0.755em},clip]{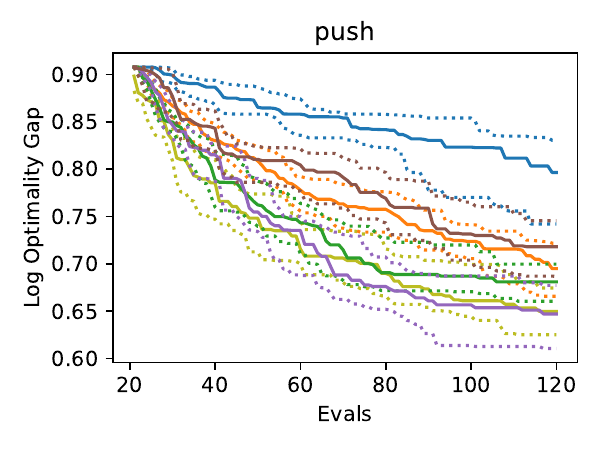}
        \\
        \hline
        \includegraphics[width=0.47\linewidth,trim={0.50em 1.25em 1.05em 0.755em},clip]{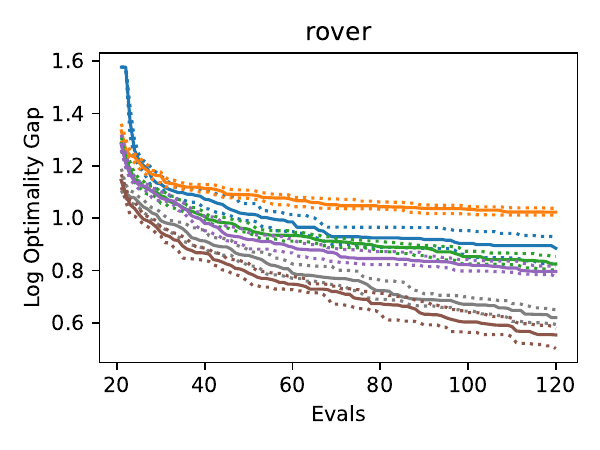}
        &
        \includegraphics[width=0.47\linewidth,trim={0.50em 1.25em 1.05em 0.755em},clip]{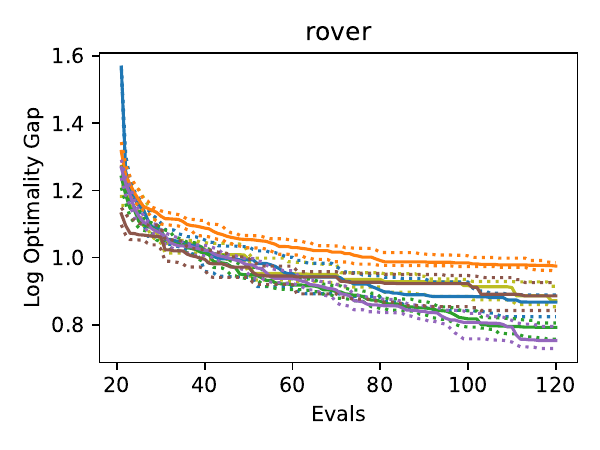}
        \\ 
        \hline
        \multicolumn{2}{|c|}{\includegraphics[width=0.99\linewidth]{images/legend.pdf}} \\
        \hline
    \end{tabular}
    
    \caption{\textbf{Results for Pygame test functions}; see Figure \ref{fig:ackley} caption.}
    \label{fig:toy}
\end{figure*}

\begin{figure*}
    \centering
    \begin{tabular}{|c|c|}
        \hline
        \textbf{Batch Size 1} & \textbf{Batch Size 10} \\
        \hline
        \includegraphics[width=0.47\linewidth,trim={0.50em 1.25em 1.05em 0.755em},clip]{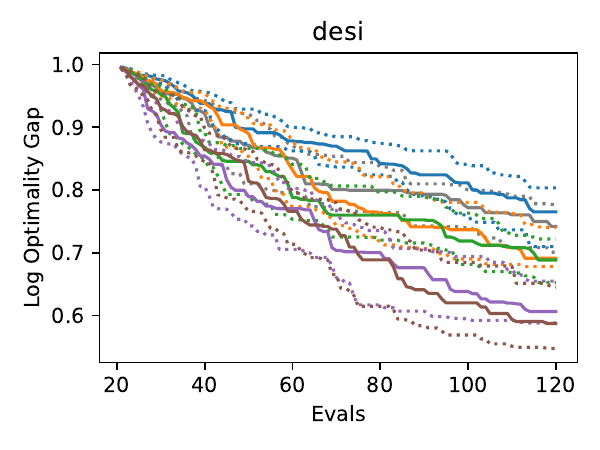}
        &
        \includegraphics[width=0.47\linewidth,trim={0.50em 1.25em 1.05em 0.755em},clip]{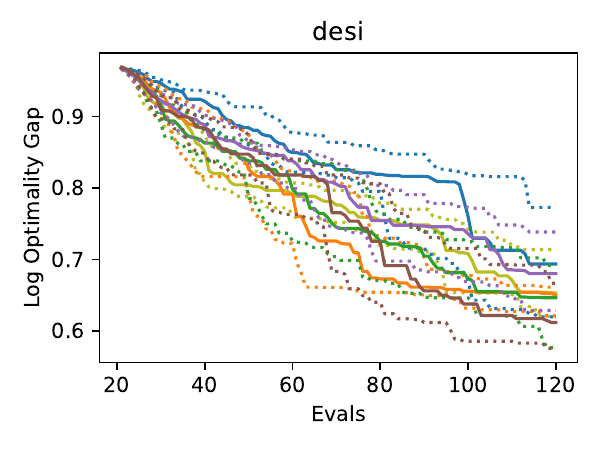}
        \\ 
        \hline
        \includegraphics[width=0.47\linewidth,trim={0.50em 1.25em 1.05em 0.755em},clip]{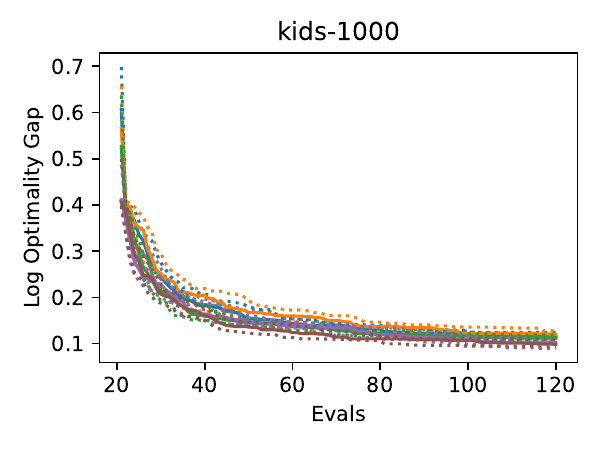}
        &
        \includegraphics[width=0.47\linewidth,trim={0.50em 1.25em 1.05em 0.755em},clip]{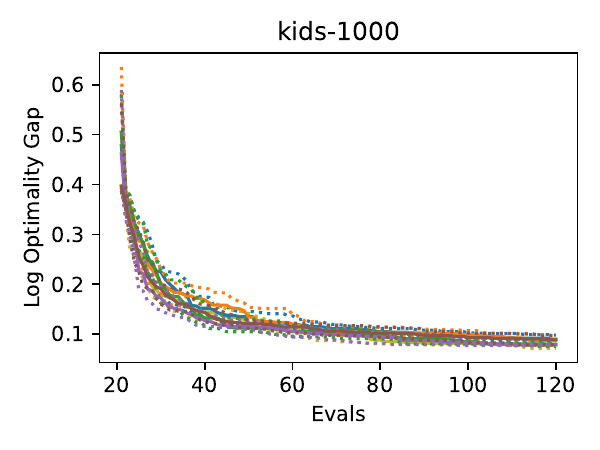}
        \\
        \hline
        \includegraphics[width=0.47\linewidth,trim={0.50em 1.25em 1.05em 0.755em},clip]{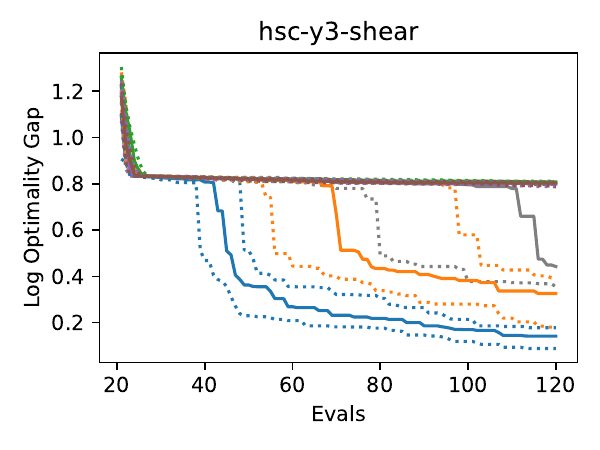}
        &
        \includegraphics[width=0.47\linewidth,trim={0.50em 1.25em 1.05em 0.755em},clip]{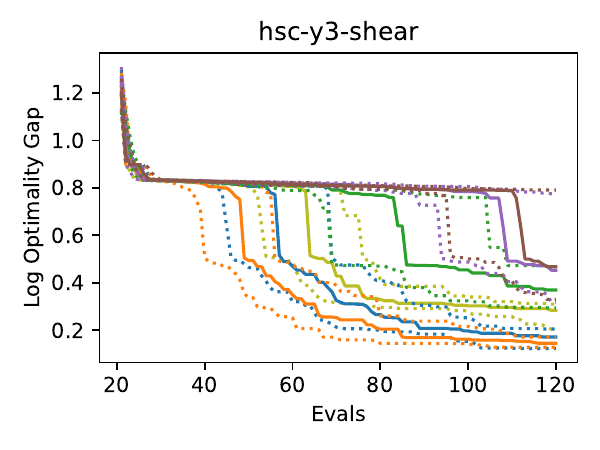}
        \\ 
        \hline
        \multicolumn{2}{|c|}{\includegraphics[width=0.99\linewidth]{images/legend.pdf}} \\
        \hline
    \end{tabular}
    
    \caption{\textbf{Results for astrophysics test functions}; see Figure \ref{fig:ackley} caption.}
    \label{fig:astro}
\end{figure*}

\begin{figure*}
    \centering
    \begin{tabular}{|c|}
        \hline
        \textbf{Batch Size 1}  \\
        \hline
        \includegraphics[width=0.99\linewidth,trim={1em 0em 1em 0em},clip]{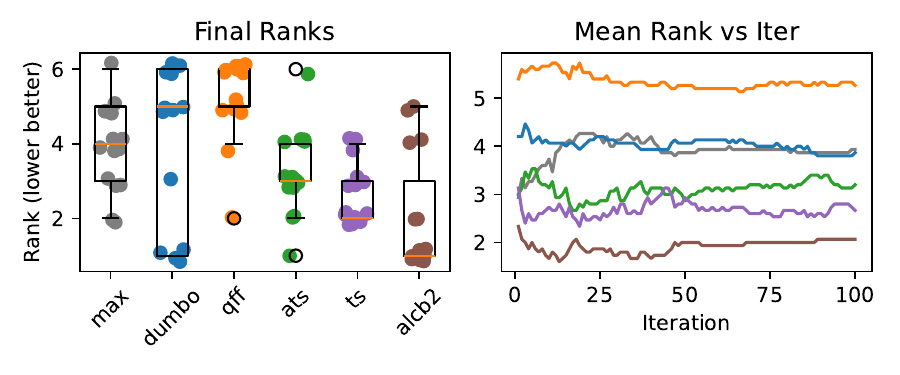}
        \\
        \hline
        \textbf{Batch Size 10} \\
        \hline
        \includegraphics[width=0.99\linewidth,trim={1em 0em 1em 0em},clip]{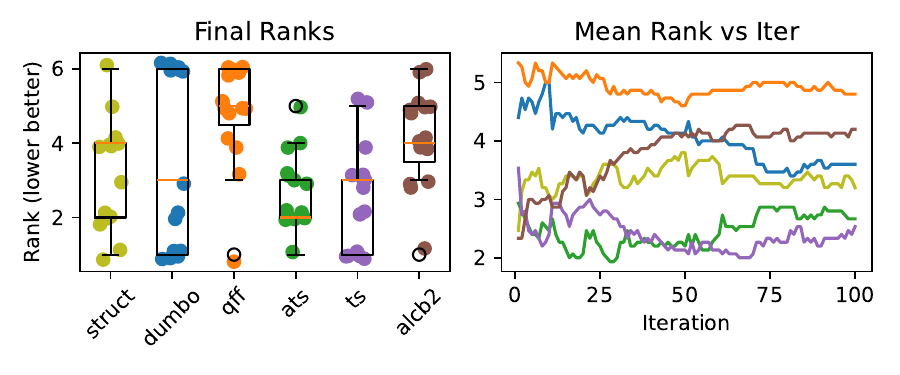}\\
        \hline
    \end{tabular}
    \caption{\textbf{Rank-Aggregated Results.}
    Shown are the ranks of each optimization method's median performance.
    In each panel, the left plot gives the performance at the end of the optimization period for each function.
    The right plot shows the mean rank averaged across functions versus sequential iteration count.
    Colors are shared between boxplot and line plot.
    }
    \label{fig:agg}
\end{figure*}

\section{Numerical Experiments}
\label{sec:results}

We are now prepared to quantitatively evaluate the impact of accounting for bilateral uncertainty in Thompson sampling.
We begin by describing our test functions and benchmark comparators.
Next we give our experimental design before presenting a summary of our results.

\subsection{Test Function Details}
\label{sec:funcs}

We now describe in detail the 15 test functions used, taking care to note differences from their previous appearances in BO.

\subsubsection{Synthetic Functions}
\label{sec:app_synthetic}
We used three common synthetic test functions, though we used them in a slightly nonstandard way.
First, we define one dimensional components for each of our test functions.
The Ackley function is defined as:
\begin{equation}
    g_a(x) = -a\exp(-b|x|)-\exp(\cos(c*x))+a+e \,,
\end{equation}
and is evaluated within [-32.768,32.768]. We used $a = 20,b=0.2, c=0.1\pi$. 

In 1D, the Levy function is defined as:
\begin{equation}
    g_l(x) = \sin(\pi w(x))^2 + (w(x)-1)^2(1+(\sin(2\pi w(x))^2))\,,
\end{equation}
with $w(x) = 1+(x-1)/4$ for $x \in [-10,10]$.

The Rastrigin function is defined as 
\begin{equation}
     g_r(x) = x^2 - 2\cos(2\pi x)      \,,
\end{equation}    
for $x\in[-\frac{3}{2},\frac{3}{2}]$.
After transforming the domain, each of these functions is parameterized such that their global optimum is at $0.5$.
To avoid benefiting acquisition methods which arbitrarily prioritize this point, we next randomly perturbed each component by a uniform variate on $[-0.5,0.5]$ such that the new optimum was uniformly distributed within $[0,1]$, that is, we sampled $u\sim U([-0.5,0.5])$ and evaluated $g_a(x-u), g_l(x-u), g_r(x-u)$ instead of simply $g_a(x), g_l(x), g_r(x)$. 
Next, we assembled $P$ many random copies of these into a $P$-dimensional function via summation.
Therefore, we defined the Ackley black-box used in experiments as:
\begin{equation}
    f_a(\x;\mathbf{u}) = \sum_{p=1}^P g_a(x_p-u_p) \,,
\end{equation}
where $\mathbf{u} = [u_1, \ldots, u_P]$ contains iid uniform $[-0.5,0.5]$ variates.
In each replication of the experiment, we sample a new vector $\mathbf{u}$ randomly such that algorithms cannot rely on the global optimum being in any particular part of the space.
We defined our high dimensional Levy and Rastrigin functions similarly.

\subsubsection{Pygame Functions}
\label{sec:app_game}

We now describe the three functions built on the python \texttt{gym} package.
We use the 14 dimensional ``push" function, obtained from the repository associated with \cite{wang2018batched}.
We also use the ``lunar landing" function, obtained from the repository associated with \cite{eriksson2019scalable}.
We changed the maximum number of steps to 100 for a faster simulation.
Finally, we used the 60 dimensional ``rover" function, also from \cite{wang2018batched}. 
We noticed that this function would fail if too many duplicate inputs on the boundaries were requested.
We handled this situation by returning a constant value of $40$ in such cases.

\subsubsection{Cosmological Inference Test Functions}
\label{sec:app_cosmo}

Cosmology and BO have long been studied together.
In this article, we created a wrapper for three of the cosmological test problems provided with CosmoSIS \cite{zuntz2015cosmosis}, software which abstracts many of the steps required for doing inference on the values of parameters in a cosmological model. 
The first example problem we used we term \texttt{desi} \citep{adame2024desi} which uses the file \texttt{examples/desi.ini}, the second \texttt{kids-1000} \citep{asgari2021kids} which uses the file \texttt{examples/kids-1000.ini}, and the third \texttt{hsc-y3-shear} \citep{dalal2023hyper}, which uses the file \texttt{examples/hsc-y3-shear.ini}, all obtained from \url{https://github.com/joezuntz/cosmosis-standard-library}.
Though similar problems have been reported in the BO literature in the past, we are not aware of previous BO work on these recently gathered datasets.

We used the default parameter files to determine which parameters to use and what bounds to use. 
For some parameters, the prior distribution imposed bounds, which we used for BO.
For other, the priors did not impose bounds.
We programatically searched for other problems using bounds on parameters of the same name.
If we did not find any, we left the parameters fixed.
Ultimately, we ended up with 7 parameters for \texttt{desi}, 14 for \texttt{kids-1000} and 25 for \texttt{hsc-y3-shear}.
The parameter names and bounds we used are given in Tables \ref{tab:desi}, \ref{tab:kids} and \ref{tab:hsc} of the Appendix.

\subsection{Experimental Design}

We conduct a large numerical study to evaluate the different acquisition methods.
We implement joint Thompson Sampling (Section \ref{sec:joint}; \texttt{ts}), its additive approximation (Section \ref{sec:indep}; \texttt{ats}) as well as the Additive Lower Confidence Bound method with $\beta$, the standard deviation multiplier, set to $2$ (\citet{kandasamy2015high}; \texttt{alcb2}) using custom JAX \citep{jax2018github} code.
This allows us to answer our primary research question by comparing the performance of \texttt{ts} and \texttt{ats}, and provides useful context by comparing against an alternative acquisition strategy, namely \texttt{alcb2}.
In order to to validate our custom framework relative to incumbent ones, we also compare to a more sophisticated LCB strategy (\citet{bardou2023relaxing}; \texttt{dumbo}), a Thompson Sampling + Fourier Feature approach (\citet{mutny2018efficient}; \texttt{qff}), and two entropy-based methods, namely \citet{wang2017max}, denoted \texttt{max}, and \citet{wang2017batched}, denoted \texttt{struct}, availing ourselves of the open source repositories associated with those articles.


We run two main experiments, one involving acquiring a single function evaluation at a time, and one with batched acquisitions of size $10$.
For the Thompson Sampling based methods (\texttt{ts}, \texttt{ats}, and \texttt{qff}), batch acquisition can be straightforwardly conducted by simply calling the methods multiple times with different random seeds \citep{kandasamy2018parallelised}.
Though not designed for batch acquisition, we converted the LCB methods (\texttt{alcb2} and \texttt{dumbo}) to this purpose using hallucinations \citep{ginsbourger2011dealing,desautels2014parallelizing}.
As they operate on similar principles, we use \texttt{max} only for the single acquisition problem and \texttt{struct} only for the batch acquisition problem.

We ran each optimization method on 9 synthetic functions, namely the \texttt{ackley}, \texttt{levy}, and \texttt{rastrigin} functions in dimensions 2, 5 and 10 (see Appendix \ref{sec:app_synthetic}).
We also used three realistic benchmarks previously used in the BO community, namely the \texttt{lunar}, \texttt{push} and \texttt{rover} problems (see Appendix \ref{sec:app_game}).
Cosmological test problems have also previously been used by the BO community (e.g. \citep{wang2018batched,eriksson2019scalable}).
Here, we investigated three recent cosmological learning problems from the CosmoSIS \citep{zuntz2015cosmosis} package, called \texttt{desi}, \texttt{kids-1000}, and \texttt{hsc-y3-shear} (see Appendix \ref{sec:app_cosmo}).

For all problems, we initialized with 20 design points sampled from a uniform distribution over the input space.
Subsequently, the methods chose 100 points sequentially.
Each experiment was replicated 100 times with different random seeds, and aggregated results are displayed.

\subsection{Benchmark Comparator Details}
\label{sec:comparators}

From the repositories of \citet{wang2017max}/\texttt{max}\footnote{\url{https://github.com/zi-w/Max-value-Entropy-Search}} and \citet{wang2017batched}/\texttt{struct}\footnote{\url{https://github.com/zi-w/Structural-Kernel-Learning-for-HDBBO}}, we adapted the scripts \texttt{example\_addgp.m} (in each repo) using Matlab's Python interface to call our functions.
Subsequently, we wrote a Python script to call these Matlab scripts using a subprocess.
These repositories relied on a custom implementation matrix inversion using the cholesky decomposition which we could not get to compile using our version of Matlab and compiler toolchain. 
Therefore, we replaced these calls with the standard Matlab \texttt{inv} calls.
Since we had comparatively small budgets, this hopefully did not introduce significant numerical instability.

For \texttt{dumbo}, we use the Python repository associated with \cite{bardou2023relaxing}, which we did not need to make any modifications to in order to incorporate in our study.

For \texttt{qff}, since the Python repository associated with \cite{mutny2018efficient} only contains the Quadrature Fourier Features, and not the code used to conduct their experiments, we wrote our own implementation.
The Fourier Feature framework does not provide a means of estimating kernel parameters; \cite{mutny2018efficient} resolved this by manually setting different parameters for different functions.
We simply set the kernel hyperparameter at $0.1$ for all experiments, which seemed to be reasonable.
We used QFF by initializing with a grid search and refining with L-BFGS-B.

In our own code, we estimated kernel hyperparameters by maximizing the marginal likelihood using the Adam \cite{kingma2014adam} algorithm with a step size of $10^{-4}$ and 1,000 iterations.
For the batch size of 1 case, we reoptimized after every 10 acquisition, and we did so after every batch in the batch size 10 case.
Before optimizing a kernel, we sampled a random additive structure with subfunctions of dimension at most 5 using the random measure on partitions, borrowing the code of \citet{bardou2023relaxing} to do so.
We sampled 500 points uniformly at random within the design space at which we evaluated either TS or LCB.

\subsection{Results}

We record the Best Observed Value (BOV) as a function of the number of sequential observations. 
Figures \ref{fig:ackley}-\ref{fig:astro} presents the median difference by competitor for most test functions.
Figure \ref{fig:agg} shows aggregated, high level results, where in order to account for the differing scales of different functions, we report rank-based statistics.

Regarding our primary research question concerning the advantage of joint versus marginal Thompson Sampling, we see that the results are mixed.
In the batch size 1 case, the joint method does appear to have a better rank across most functions and iterations than the approximate method, but it does not appear to be a large difference. 
And in the batch size 10 case, which is best seems to depend on iteration count and the specific function.
Further figures given in Appendix \ref{sec:bilateral} show that on most individual problems, the lead of \texttt{ts} is not so great. 
A notable exception is the \texttt{desi} problem, with a batch size of 1, where \texttt{ts} does seem to have a significant advantage; however, this advantage is not present on a batch size of 10, and the \texttt{alcb2} method anyways does slightly better in the batch size 1 case.
\texttt{alcb2} does seem to generally be superior to the Thompson Sampling based methods in the batch size 1 case, but hallucinations are not enough to make it competitive for batch size 10.
In sanity-checking our own implementation against existing ones \texttt{max}, \texttt{struct}, and \texttt{dumbo}, we find confidence in our implementation of BO as it routinely matches or outperforms then, and conclude that it can serve as a fair comparison between \texttt{ts}, \texttt{ats}, and \texttt{alcb2}.
Appendix \ref{sec:boxes} shows the entire distribution of results for the BOV at final iteration.

\section{Discussion}
\label{sec:conclusion}

\paragraph{Summary:}
In this article, we showed how to efficiently conduct Thompson Sampling on Additive Gaussian Processes.
In 1,500 total trials of 6 methods across 15 test functions, we showed that in the context of Bayesian optimization, taking bilateral uncertainty into account provided a small advantage in the sequential acquisition case and that there was no consistent advantage in the batch size 10 case.

\paragraph{Conclusions:}
Our empirical results complement existing theory in establishing that omitting bilateral uncertainty does significantly not sacrifice performance.
Taken on their own, these results suggests that future research in additive BO can focus on other aspects of the BO process such as model fidelity, and that in proposing new acquisition functions, we should not hesitate to propose them directly on the subfunctions.
But it is not obvious that there could not be some pathological example where the bilateral uncertainty is key in small samples.
Construction of such a function or a proof ruling it out is thus of interest.

\paragraph{Future Work:}
It will be of interest to study how important bilateral uncertainty is for other uses of Thompson Sampling aside from BO, such as reinforcement learning.
Also, in order to scale to larger problems, it will be essential to incorporate recent proposals for large-scale sampling of GPs such as \citet{wilson2020efficiently} and \citet{vakili2021scalable}.

\section*{Acknowledgments}

I thank Micka\"el Binois for introducing me to this area and Jake Gardner for a helpful email discussion early in the conception of this project; of course any mistakes and oversights are entirely my own. 
I gratefully acknowledge funding from the NSF and NGA via DMS\#2428033 as well as the NSF and AFOSR via DMS\#2529277, as well as computing resources from the The Massachusetts Green High Performance Computing Center.

\bibliography{main}


\begin{thebibliography}{32}
\ifx \bisbn   \undefined \def \bisbn  #1{ISBN #1}\fi
\ifx \binits  \undefined \def \binits#1{#1}\fi
\ifx \bauthor  \undefined \def \bauthor#1{#1}\fi
\ifx \batitle  \undefined \def \batitle#1{#1}\fi
\ifx \bjtitle  \undefined \def \bjtitle#1{#1}\fi
\ifx \bvolume  \undefined \def \bvolume#1{\textbf{#1}}\fi
\ifx \byear  \undefined \def \byear#1{#1}\fi
\ifx \bissue  \undefined \def \bissue#1{#1}\fi
\ifx \bfpage  \undefined \def \bfpage#1{#1}\fi
\ifx \blpage  \undefined \def \blpage #1{#1}\fi
\ifx \burl  \undefined \def \burl#1{\textsf{#1}}\fi
\ifx \doiurl  \undefined \def \doiurl#1{\url{https://doi.org/#1}}\fi
\ifx \betal  \undefined \def \betal{\textit{et al.}}\fi
\ifx \binstitute  \undefined \def \binstitute#1{#1}\fi
\ifx \binstitutionaled  \undefined \def \binstitutionaled#1{#1}\fi
\ifx \bctitle  \undefined \def \bctitle#1{#1}\fi
\ifx \beditor  \undefined \def \beditor#1{#1}\fi
\ifx \bpublisher  \undefined \def \bpublisher#1{#1}\fi
\ifx \bbtitle  \undefined \def \bbtitle#1{#1}\fi
\ifx \bedition  \undefined \def \bedition#1{#1}\fi
\ifx \bseriesno  \undefined \def \bseriesno#1{#1}\fi
\ifx \blocation  \undefined \def \blocation#1{#1}\fi
\ifx \bsertitle  \undefined \def \bsertitle#1{#1}\fi
\ifx \bsnm \undefined \def \bsnm#1{#1}\fi
\ifx \bsuffix \undefined \def \bsuffix#1{#1}\fi
\ifx \bparticle \undefined \def \bparticle#1{#1}\fi
\ifx \barticle \undefined \def \barticle#1{#1}\fi
\bibcommenthead
\ifx \bconfdate \undefined \def \bconfdate #1{#1}\fi
\ifx \botherref \undefined \def \botherref #1{#1}\fi
\ifx \url \undefined \def \url#1{\textsf{#1}}\fi
\ifx \bchapter \undefined \def \bchapter#1{#1}\fi
\ifx \bbook \undefined \def \bbook#1{#1}\fi
\ifx \bcomment \undefined \def \bcomment#1{#1}\fi
\ifx \oauthor \undefined \def \oauthor#1{#1}\fi
\ifx \citeauthoryear \undefined \def \citeauthoryear#1{#1}\fi
\ifx \endbibitem  \undefined \def \endbibitem {}\fi
\ifx \bconflocation  \undefined \def \bconflocation#1{#1}\fi
\ifx \arxivurl  \undefined \def \arxivurl#1{\textsf{#1}}\fi
\csname PreBibitemsHook\endcsname

\bibitem[\protect\citeauthoryear{Garnett}{2023}]{garnett_bayesoptbook_2023}
\begin{bbook}
\bauthor{\bsnm{Garnett}, \binits{R.}}:
\bbtitle{{B}ayesian {O}ptimization}.
\bpublisher{Cambridge University Press},
\blocation{Cambridge, UK}
(\byear{2023})
\end{bbook}
\endbibitem

\bibitem[\protect\citeauthoryear{Kandasamy et~al.}{2015}]{kandasamy2015high}
\begin{bchapter}
\bauthor{\bsnm{Kandasamy}, \binits{K.}},
\bauthor{\bsnm{Schneider}, \binits{J.}},
\bauthor{\bsnm{P{\'o}czos}, \binits{B.}}:
\bctitle{High dimensional bayesian optimisation and bandits via additive models}.
In: \bbtitle{International Conference on Machine Learning},
pp. \bfpage{295}--\blpage{304}
(\byear{2015}).
\bcomment{PMLR}
\end{bchapter}
\endbibitem

\bibitem[\protect\citeauthoryear{Durrande et~al.}{2011}]{durrande2011additive}
\begin{botherref}
\oauthor{\bsnm{Durrande}, \binits{N.}},
\oauthor{\bsnm{Ginsbourger}, \binits{D.}},
\oauthor{\bsnm{Roustant}, \binits{O.}}:
Additive kernels for gaussian process modeling.
arXiv preprint arXiv:1103.4023
(2011)
\end{botherref}
\endbibitem

\bibitem[\protect\citeauthoryear{Duvenaud et~al.}{2011}]{duvenaud2011additive}
\begin{botherref}
\oauthor{\bsnm{Duvenaud}, \binits{D.K.}},
\oauthor{\bsnm{Nickisch}, \binits{H.}},
\oauthor{\bsnm{Rasmussen}, \binits{C.}}:
Additive gaussian processes.
Advances in neural information processing systems
\textbf{24}
(2011)
\end{botherref}
\endbibitem

\bibitem[\protect\citeauthoryear{Nesterov}{2013}]{nesterov2013introductory}
\begin{bbook}
\bauthor{\bsnm{Nesterov}, \binits{Y.}}:
\bbtitle{Introductory Lectures on Convex Optimization: A Basic Course}
vol. \bseriesno{87}.
\bpublisher{Springer},
\blocation{Norwell, MA}
(\byear{2013})
\end{bbook}
\endbibitem

\bibitem[\protect\citeauthoryear{Mutny and Krause}{2018}]{mutny2018efficient}
\begin{botherref}
\oauthor{\bsnm{Mutny}, \binits{M.}},
\oauthor{\bsnm{Krause}, \binits{A.}}:
Efficient high dimensional bayesian optimization with additivity and quadrature fourier features.
Advances in Neural Information Processing Systems
\textbf{31}
(2018)
\end{botherref}
\endbibitem

\bibitem[\protect\citeauthoryear{Thompson}{1933}]{thompson1933likelihood}
\begin{barticle}
\bauthor{\bsnm{Thompson}, \binits{W.R.}}:
\batitle{On the likelihood that one unknown probability exceeds another in view of the evidence of two samples}.
\bjtitle{Biometrika}
\bvolume{25}(\bissue{3-4}),
\bfpage{285}--\blpage{294}
(\byear{1933})
\end{barticle}
\endbibitem

\bibitem[\protect\citeauthoryear{Wang and Jegelka}{2017}]{wang2017max}
\begin{bchapter}
\bauthor{\bsnm{Wang}, \binits{Z.}},
\bauthor{\bsnm{Jegelka}, \binits{S.}}:
\bctitle{Max-value entropy search for efficient bayesian optimization}.
In: \bbtitle{International Conference on Machine Learning},
pp. \bfpage{3627}--\blpage{3635}
(\byear{2017}).
\bcomment{PMLR}
\end{bchapter}
\endbibitem

\bibitem[\protect\citeauthoryear{Bardou et~al.}{2023}]{bardou2023relaxing}
\begin{botherref}
\oauthor{\bsnm{Bardou}, \binits{A.}},
\oauthor{\bsnm{Thiran}, \binits{P.}},
\oauthor{\bsnm{Begin}, \binits{T.}}:
Relaxing the additivity constraints in decentralized no-regret high-dimensional bayesian optimization.
arXiv preprint arXiv:2305.19838
(2023)
\end{botherref}
\endbibitem

\bibitem[\protect\citeauthoryear{Gardner et~al.}{2017}]{gardner2017discovering}
\begin{bchapter}
\bauthor{\bsnm{Gardner}, \binits{J.}},
\bauthor{\bsnm{Guo}, \binits{C.}},
\bauthor{\bsnm{Weinberger}, \binits{K.}},
\bauthor{\bsnm{Garnett}, \binits{R.}},
\bauthor{\bsnm{Grosse}, \binits{R.}}:
\bctitle{Discovering and exploiting additive structure for bayesian optimization}.
In: \bbtitle{Artificial Intelligence and Statistics},
pp. \bfpage{1311}--\blpage{1319}
(\byear{2017}).
\bcomment{PMLR}
\end{bchapter}
\endbibitem

\bibitem[\protect\citeauthoryear{Han et~al.}{2021}]{han2021high}
\begin{bchapter}
\bauthor{\bsnm{Han}, \binits{E.}},
\bauthor{\bsnm{Arora}, \binits{I.}},
\bauthor{\bsnm{Scarlett}, \binits{J.}}:
\bctitle{High-dimensional bayesian optimization via tree-structured additive models}.
In: \bbtitle{Proceedings of the AAAI Conference on Artificial Intelligence},
vol. \bseriesno{35},
pp. \bfpage{7630}--\blpage{7638}
(\byear{2021})
\end{bchapter}
\endbibitem

\bibitem[\protect\citeauthoryear{Ziomek and Ammar}{2023}]{ziomek2023random}
\begin{bchapter}
\bauthor{\bsnm{Ziomek}, \binits{J.K.}},
\bauthor{\bsnm{Ammar}, \binits{H.B.}}:
\bctitle{Are random decompositions all we need in high dimensional bayesian optimisation?}
In: \bbtitle{International Conference on Machine Learning},
pp. \bfpage{43347}--\blpage{43368}
(\byear{2023}).
\bcomment{PMLR}
\end{bchapter}
\endbibitem

\bibitem[\protect\citeauthoryear{Wang et~al.}{2017}]{wang2017batched}
\begin{bchapter}
\bauthor{\bsnm{Wang}, \binits{Z.}},
\bauthor{\bsnm{Li}, \binits{C.}},
\bauthor{\bsnm{Jegelka}, \binits{S.}},
\bauthor{\bsnm{Kohli}, \binits{P.}}:
\bctitle{Batched high-dimensional bayesian optimization via structural kernel learning}.
In: \bbtitle{International Conference on Machine Learning},
pp. \bfpage{3656}--\blpage{3664}
(\byear{2017}).
\bcomment{PMLR}
\end{bchapter}
\endbibitem

\bibitem[\protect\citeauthoryear{Rolland et~al.}{2018}]{rolland2018high}
\begin{bchapter}
\bauthor{\bsnm{Rolland}, \binits{P.}},
\bauthor{\bsnm{Scarlett}, \binits{J.}},
\bauthor{\bsnm{Bogunovic}, \binits{I.}},
\bauthor{\bsnm{Cevher}, \binits{V.}}:
\bctitle{High-dimensional bayesian optimization via additive models with overlapping groups}.
In: \bbtitle{International Conference on Artificial Intelligence and Statistics},
pp. \bfpage{298}--\blpage{307}
(\byear{2018}).
\bcomment{PMLR}
\end{bchapter}
\endbibitem

\bibitem[\protect\citeauthoryear{Hoang et~al.}{2018}]{hoang2018decentralized}
\begin{bchapter}
\bauthor{\bsnm{Hoang}, \binits{T.N.}},
\bauthor{\bsnm{Hoang}, \binits{Q.M.}},
\bauthor{\bsnm{Ouyang}, \binits{R.}},
\bauthor{\bsnm{Low}, \binits{K.H.}}:
\bctitle{Decentralized high-dimensional bayesian optimization with factor graphs}.
In: \bbtitle{Proceedings of the AAAI Conference on Artificial Intelligence},
vol. \bseriesno{32}
(\byear{2018})
\end{bchapter}
\endbibitem

\bibitem[\protect\citeauthoryear{Lu et~al.}{2022}]{lu2022additive}
\begin{bchapter}
\bauthor{\bsnm{Lu}, \binits{X.}},
\bauthor{\bsnm{Boukouvalas}, \binits{A.}},
\bauthor{\bsnm{Hensman}, \binits{J.}}:
\bctitle{Additive gaussian processes revisited}.
In: \bbtitle{International Conference on Machine Learning},
pp. \bfpage{14358}--\blpage{14383}
(\byear{2022}).
\bcomment{PMLR}
\end{bchapter}
\endbibitem

\bibitem[\protect\citeauthoryear{Rahimi and Recht}{2007}]{rahimi2007random}
\begin{botherref}
\oauthor{\bsnm{Rahimi}, \binits{A.}},
\oauthor{\bsnm{Recht}, \binits{B.}}:
Random features for large-scale kernel machines.
Advances in neural information processing systems
\textbf{20}
(2007)
\end{botherref}
\endbibitem

\bibitem[\protect\citeauthoryear{Seeger}{2004}]{seeger2004low}
\begin{botherref}
\oauthor{\bsnm{Seeger}, \binits{M.}}:
Low rank updates for the cholesky decomposition
(2004)
\end{botherref}
\endbibitem

\bibitem[\protect\citeauthoryear{Harris et~al.}{2020}]{harris2020array}
\begin{barticle}
\bauthor{\bsnm{Harris}, \binits{C.R.}},
\bauthor{\bsnm{Millman}, \binits{K.J.}},
\bauthor{\bsnm{Walt}, \binits{S.J.}},
\bauthor{\bsnm{Gommers}, \binits{R.}},
\bauthor{\bsnm{Virtanen}, \binits{P.}},
\bauthor{\bsnm{Cournapeau}, \binits{D.}},
\bauthor{\bsnm{Wieser}, \binits{E.}},
\bauthor{\bsnm{Taylor}, \binits{J.}},
\bauthor{\bsnm{Berg}, \binits{S.}},
\bauthor{\bsnm{Smith}, \binits{N.J.}},
\bauthor{\bsnm{Kern}, \binits{R.}},
\bauthor{\bsnm{Picus}, \binits{M.}},
\bauthor{\bsnm{Hoyer}, \binits{S.}},
\bauthor{\bsnm{Kerkwijk}, \binits{M.H.}},
\bauthor{\bsnm{Brett}, \binits{M.}},
\bauthor{\bsnm{Haldane}, \binits{A.}},
\bauthor{\bsnm{R{\'{i}}o}, \binits{J.F.}},
\bauthor{\bsnm{Wiebe}, \binits{M.}},
\bauthor{\bsnm{Peterson}, \binits{P.}},
\bauthor{\bsnm{G{\'{e}}rard-Marchant}, \binits{P.}},
\bauthor{\bsnm{Sheppard}, \binits{K.}},
\bauthor{\bsnm{Reddy}, \binits{T.}},
\bauthor{\bsnm{Weckesser}, \binits{W.}},
\bauthor{\bsnm{Abbasi}, \binits{H.}},
\bauthor{\bsnm{Gohlke}, \binits{C.}},
\bauthor{\bsnm{Oliphant}, \binits{T.E.}}:
\batitle{Array programming with {NumPy}}.
\bjtitle{Nature}
\bvolume{585}(\bissue{7825}),
\bfpage{357}--\blpage{362}
(\byear{2020})
\doiurl{10.1038/s41586-020-2649-2}
\end{barticle}
\endbibitem

\bibitem[\protect\citeauthoryear{Wang et~al.}{2018}]{wang2018batched}
\begin{bchapter}
\bauthor{\bsnm{Wang}, \binits{Z.}},
\bauthor{\bsnm{Gehring}, \binits{C.}},
\bauthor{\bsnm{Kohli}, \binits{P.}},
\bauthor{\bsnm{Jegelka}, \binits{S.}}:
\bctitle{Batched large-scale bayesian optimization in high-dimensional spaces}.
In: \bbtitle{International Conference on Artificial Intelligence and Statistics},
pp. \bfpage{745}--\blpage{754}
(\byear{2018}).
\bcomment{PMLR}
\end{bchapter}
\endbibitem

\bibitem[\protect\citeauthoryear{Eriksson et~al.}{2019}]{eriksson2019scalable}
\begin{botherref}
\oauthor{\bsnm{Eriksson}, \binits{D.}},
\oauthor{\bsnm{Pearce}, \binits{M.}},
\oauthor{\bsnm{Gardner}, \binits{J.}},
\oauthor{\bsnm{Turner}, \binits{R.D.}},
\oauthor{\bsnm{Poloczek}, \binits{M.}}:
Scalable global optimization via local bayesian optimization.
Advances in neural information processing systems
\textbf{32}
(2019)
\end{botherref}
\endbibitem

\bibitem[\protect\citeauthoryear{Zuntz et~al.}{2015}]{zuntz2015cosmosis}
\begin{barticle}
\bauthor{\bsnm{Zuntz}, \binits{J.}},
\bauthor{\bsnm{Paterno}, \binits{M.}},
\bauthor{\bsnm{Jennings}, \binits{E.}},
\bauthor{\bsnm{Rudd}, \binits{D.}},
\bauthor{\bsnm{Manzotti}, \binits{A.}},
\bauthor{\bsnm{Dodelson}, \binits{S.}},
\bauthor{\bsnm{Bridle}, \binits{S.}},
\bauthor{\bsnm{Sehrish}, \binits{S.}},
\bauthor{\bsnm{Kowalkowski}, \binits{J.}}:
\batitle{Cosmosis: Modular cosmological parameter estimation}.
\bjtitle{Astronomy and Computing}
\bvolume{12},
\bfpage{45}--\blpage{59}
(\byear{2015})
\end{barticle}
\endbibitem

\bibitem[\protect\citeauthoryear{Adame et~al.}{2024}]{adame2024desi}
\begin{botherref}
\oauthor{\bsnm{Adame}, \binits{A.}},
\oauthor{\bsnm{Aguilar}, \binits{J.}},
\oauthor{\bsnm{Ahlen}, \binits{S.}},
\oauthor{\bsnm{Alam}, \binits{S.}},
\oauthor{\bsnm{Alexander}, \binits{D.}},
\oauthor{\bsnm{Alvarez}, \binits{M.}},
\oauthor{\bsnm{Alves}, \binits{O.}},
\oauthor{\bsnm{Anand}, \binits{A.}},
\oauthor{\bsnm{Andrade}, \binits{U.}},
\oauthor{\bsnm{Armengaud}, \binits{E.}}, et al.:
Desi 2024 vi: Cosmological constraints from the measurements of baryon acoustic oscillations.
arXiv preprint arXiv:2404.03002
(2024)
\end{botherref}
\endbibitem

\bibitem[\protect\citeauthoryear{Asgari et~al.}{2021}]{asgari2021kids}
\begin{barticle}
\bauthor{\bsnm{Asgari}, \binits{M.}},
\bauthor{\bsnm{Lin}, \binits{C.-A.}},
\bauthor{\bsnm{Joachimi}, \binits{B.}},
\bauthor{\bsnm{Giblin}, \binits{B.}},
\bauthor{\bsnm{Heymans}, \binits{C.}},
\bauthor{\bsnm{Hildebrandt}, \binits{H.}},
\bauthor{\bsnm{Kannawadi}, \binits{A.}},
\bauthor{\bsnm{St{\"o}lzner}, \binits{B.}},
\bauthor{\bsnm{Tr{\"o}ster}, \binits{T.}},
\bauthor{\bsnm{Busch}, \binits{J.L.}}, \betal:
\batitle{Kids-1000 cosmology: Cosmic shear constraints and comparison between two point statistics}.
\bjtitle{Astronomy \& Astrophysics}
\bvolume{645},
\bfpage{104}
(\byear{2021})
\end{barticle}
\endbibitem

\bibitem[\protect\citeauthoryear{Dalal et~al.}{2023}]{dalal2023hyper}
\begin{barticle}
\bauthor{\bsnm{Dalal}, \binits{R.}},
\bauthor{\bsnm{Li}, \binits{X.}},
\bauthor{\bsnm{Nicola}, \binits{A.}},
\bauthor{\bsnm{Zuntz}, \binits{J.}},
\bauthor{\bsnm{Strauss}, \binits{M.A.}},
\bauthor{\bsnm{Sugiyama}, \binits{S.}},
\bauthor{\bsnm{Zhang}, \binits{T.}},
\bauthor{\bsnm{Rau}, \binits{M.M.}},
\bauthor{\bsnm{Mandelbaum}, \binits{R.}},
\bauthor{\bsnm{Takada}, \binits{M.}}, \betal:
\batitle{Hyper suprime-cam year 3 results: Cosmology from cosmic shear power spectra}.
\bjtitle{Physical Review D}
\bvolume{108}(\bissue{12}),
\bfpage{123519}
(\byear{2023})
\end{barticle}
\endbibitem

\bibitem[\protect\citeauthoryear{Bradbury et~al.}{2018}]{jax2018github}
\begin{botherref}
\oauthor{\bsnm{Bradbury}, \binits{J.}},
\oauthor{\bsnm{Frostig}, \binits{R.}},
\oauthor{\bsnm{Hawkins}, \binits{P.}},
\oauthor{\bsnm{Johnson}, \binits{M.J.}},
\oauthor{\bsnm{Leary}, \binits{C.}},
\oauthor{\bsnm{Maclaurin}, \binits{D.}},
\oauthor{\bsnm{Necula}, \binits{G.}},
\oauthor{\bsnm{Paszke}, \binits{A.}},
\oauthor{\bsnm{Vander{P}las}, \binits{J.}},
\oauthor{\bsnm{Wanderman-{M}ilne}, \binits{S.}},
\oauthor{\bsnm{Zhang}, \binits{Q.}}:
{JAX}: Composable Transformations of {P}ython+{N}um{P}y programs.
\url{http://github.com/jax-ml/jax}
\end{botherref}
\endbibitem

\bibitem[\protect\citeauthoryear{Kandasamy et~al.}{2018}]{kandasamy2018parallelised}
\begin{bchapter}
\bauthor{\bsnm{Kandasamy}, \binits{K.}},
\bauthor{\bsnm{Krishnamurthy}, \binits{A.}},
\bauthor{\bsnm{Schneider}, \binits{J.}},
\bauthor{\bsnm{P{\'o}czos}, \binits{B.}}:
\bctitle{Parallelised bayesian optimisation via thompson sampling}.
In: \bbtitle{International Conference on Artificial Intelligence and Statistics},
pp. \bfpage{133}--\blpage{142}
(\byear{2018}).
\bcomment{PMLR}
\end{bchapter}
\endbibitem

\bibitem[\protect\citeauthoryear{Ginsbourger et~al.}{2011}]{ginsbourger2011dealing}
\begin{botherref}
\oauthor{\bsnm{Ginsbourger}, \binits{D.}},
\oauthor{\bsnm{Janusevskis}, \binits{J.}},
\oauthor{\bsnm{Le~Riche}, \binits{R.}}:
Dealing with asynchronicity in parallel gaussian process based global optimization.
PhD thesis,
Mines Saint-Etienne
(2011)
\end{botherref}
\endbibitem

\bibitem[\protect\citeauthoryear{Desautels et~al.}{2014}]{desautels2014parallelizing}
\begin{barticle}
\bauthor{\bsnm{Desautels}, \binits{T.}},
\bauthor{\bsnm{Krause}, \binits{A.}},
\bauthor{\bsnm{Burdick}, \binits{J.W.}}:
\batitle{Parallelizing exploration-exploitation tradeoffs in gaussian process bandit optimization.}
\bjtitle{J. Mach. Learn. Res.}
\bvolume{15}(\bissue{1}),
\bfpage{3873}--\blpage{3923}
(\byear{2014})
\end{barticle}
\endbibitem

\bibitem[\protect\citeauthoryear{Kingma and Ba}{2014}]{kingma2014adam}
\begin{botherref}
\oauthor{\bsnm{Kingma}, \binits{D.P.}},
\oauthor{\bsnm{Ba}, \binits{J.}}:
Adam: A method for stochastic optimization.
arXiv e-prints,
1412
(2014)
\end{botherref}
\endbibitem

\bibitem[\protect\citeauthoryear{Wilson et~al.}{2020}]{wilson2020efficiently}
\begin{bchapter}
\bauthor{\bsnm{Wilson}, \binits{J.}},
\bauthor{\bsnm{Borovitskiy}, \binits{V.}},
\bauthor{\bsnm{Terenin}, \binits{A.}},
\bauthor{\bsnm{Mostowsky}, \binits{P.}},
\bauthor{\bsnm{Deisenroth}, \binits{M.}}:
\bctitle{Efficiently sampling functions from gaussian process posteriors}.
In: \bbtitle{International Conference on Machine Learning},
pp. \bfpage{10292}--\blpage{10302}
(\byear{2020}).
\bcomment{PMLR}
\end{bchapter}
\endbibitem

\bibitem[\protect\citeauthoryear{Vakili et~al.}{2021}]{vakili2021scalable}
\begin{barticle}
\bauthor{\bsnm{Vakili}, \binits{S.}},
\bauthor{\bsnm{Moss}, \binits{H.}},
\bauthor{\bsnm{Artemev}, \binits{A.}},
\bauthor{\bsnm{Dutordoir}, \binits{V.}},
\bauthor{\bsnm{Picheny}, \binits{V.}}:
\batitle{Scalable thompson sampling using sparse gaussian process models}.
\bjtitle{Advances in neural information processing systems}
\bvolume{34},
\bfpage{5631}--\blpage{5643}
(\byear{2021})
\end{barticle}
\endbibitem

\end{thebibliography}

\newpage

\backmatter

\appendix

\section{Numpy Implementation}

This section contains a Numpy implementation of the joint Thompson Sampler of Section \ref{sec:joint}.
Note that we iterate from $M$ to $1$ rather than $1$ to $M$ as in the main text so that we can use the \texttt{cumsum} command without having to sort.

\label{sec:numpy}
\begin{lstlisting}[frame=tb,
  language=Python,
  aboveskip=3mm,
  belowskip=3mm,
  showstringspaces=false,
  columns=flexible,
  basicstyle={\small\ttfamily},
  numbers=none,
  numberstyle=\tiny\color{gray},
  keywordstyle=\color{blue},
  commentstyle=\color{purple},
  stringstyle=\color{mauve},
  breaklines=true,
  breakatwhitespace=true,
  tabsize=3,
  caption= {Numpy implementation of Batch TS. 
  \texttt{KXX[m,:,:]} yields $\K_m(\P_m\X,\P_m\X)$,
  \texttt{KZX[m,:,:]} yields $\K_m(\Z_m,\P_m\X)$,
  and \texttt{KZZ[m,:,:]} yields $\K_m(\Z_m,\Z_m)$.
  }
  ]
import numpy as np
from numpy.linalg import solve, matrix_transpose as mt, cholesky
from numpy.random import normal

# Sample Stochastic Part
KXX_cs = np.cumsum(KXX,axis=0)
KsiK = solve(KXX_cs+tau2*np.eye(N), mt(KZX))
SIGMA_eff = KZZ_all - KZX @ KsiK
L_eff = cholesky(SIGMA_eff + eps_chol*I)
cov_part = L_eff @ normal(size=[M,di,B])

# Adjust with mean part
for m in range(M-1,-1,-1):
    pred =np.sum(eff_Z[:,(m+1):,:],axis=1)
    resid = y_use[:,None]-pred
    mu_fp = mt(KsiK[m,:,:]) @ resid
    eff_all = cov_part[m,:,:] + mu_fp
    eff_ZZ[:,m,:] = eff_all[:NN,:]
    eff_Z[:,m,:] = eff_all[NN:,:]


\end{lstlisting}

\section{Additional Numerical Experiment Details}
\label{sec:num_app}

We now provide additional details of the numerical results. 
First, we provide the specific parameters and bounds used in the cosmology experiments in Tables \ref{tab:desi}, \ref{tab:kids} and \ref{tab:hsc}.
Next, we present boxplots giving the full distribution of estimates.

\begin{table}
\centering
\begin{tabular}{|llll|}
\hline
\multicolumn{4}{|c|}{Problem \texttt{desi} Parameters}\\
\hline
Name  & Lower Bound & Upper Bound & Type \\
\hline
h0 & 0.600000 & 0.800000 & cosmological\_parameters \\
ombh2 & 0.020000 & 0.025000 & cosmological\_parameters \\
omega\_m & 0.01 & 0.99 & cosmological\_parameters \\
rdh & 10.0 & 1000.0 & cosmological\_parameters \\
n\_s & 0.860000 & 1.060000 & cosmological\_parameters \\
omega\_k & -0.300000 & 0.300000 & cosmological\_parameters \\
tau & 0.055000 & 0.092500 & cosmological\_parameters \\
\bottomrule
\end{tabular}
\caption{Parameter bounds for 7 dimensional \texttt{desi} problem.}
\label{tab:desi}
\end{table}

\begin{table}
\centering
\begin{tabular}{|llll|}
\hline
\multicolumn{4}{|c|}{Problem \texttt{kids-1000} Parameters}\\
\hline
Name  & Lower Bound & Upper Bound & Type \\
\hline
uncorr\_bias\_1 & -5.0 & 5.0 & nofz\_shifts\_kids \\
uncorr\_bias\_2 & -5.0 & 5.0 & nofz\_shifts\_kids \\
uncorr\_bias\_3 & -5.0 & 5.0 & nofz\_shifts\_kids \\
uncorr\_bias\_4 & -5.0 & 5.0 & nofz\_shifts\_kids \\
uncorr\_bias\_5 & -5.0 & 5.0 & nofz\_shifts\_kids \\
a1 & -5.0 & 5.0 & intrinsic\_alignment\_parameters \\
alpha1 & -5.0 & 5.0 & intrinsic\_alignment\_parameters \\
logt\_agn & 7.3 & 8.0 & halo\_model\_parameters \\
omch2 & 0.051 & 0.255 & cosmological\_parameters \\
ombh2 & 0.019 & 0.026 & cosmological\_parameters \\
h0 & 0.64 & 0.82 & cosmological\_parameters \\
n\_s & 0.84 & 1.1 & cosmological\_parameters \\
s\_8 & 0.1 & 1.3 & cosmological\_parameters \\
mnu & 0.055 & 0.6 & cosmological\_parameters \\
\bottomrule
\end{tabular}
\caption{Parameter bounds for 14 dimensional \texttt{kids-1000} problem.}
\label{tab:kids}
\end{table}

\begin{table}
\centering
\begin{tabular}{|llll|}
\hline
\multicolumn{4}{|c|}{Problem \texttt{hsc-y3-shear} Parameters}\\
\hline
Name  & Lower Bound & Upper Bound & Type \\
\hline
a\_s & 0.5e-9 & 10e-09 & cosmological\_parameters \\
omega\_m & 0.10 & 0.70 & cosmological\_parameters \\
n\_s & 0.87 & 1.07 & cosmological\_parameters \\
h0 & 0.62 & 0.80 & cosmological\_parameters \\
ombh2 & 0.020 & 0.025 & cosmological\_parameters \\
omega\_k & -0.300000 & 0.300000 & cosmological\_parameters \\
tau & 0.055000 & 0.092500 & cosmological\_parameters \\
a & 2.0 & 3.13 & halo\_model\_parameters \\
a1 & -6.0 & 6.0 & intrinsic\_alignment\_parameters \\
a2 & -6.0 & 6.0 & intrinsic\_alignment\_parameters \\
alpha1 & -6.0 & 6.0 & intrinsic\_alignment\_parameters \\
alpha2 & -6.0 & 6.0 & intrinsic\_alignment\_parameters \\
bias\_ta & 0.0 & 2.0 & intrinsic\_alignment\_parameters \\
bias\_1 & -1.0 & 1.0 & wl\_photoz\_errors \\
bias\_2 & -1.0 & 1.0 & wl\_photoz\_errors \\
bias\_3 & -1.0 & 1.0 & wl\_photoz\_errors \\
bias\_4 & -1.0 & 1.0 & wl\_photoz\_errors \\
m1 & -0.1 & 0.1 & shear\_calibration\_parameters \\
m2 & -0.1 & 0.1 & shear\_calibration\_parameters \\
m3 & -0.1 & 0.1 & shear\_calibration\_parameters \\
m4 & -0.1 & 0.1 & shear\_calibration\_parameters \\
psf\_alpha2 & -5.0 & 5.0 & psf\_parameters \\
psf\_beta2 & -5.0 & 5.0 & psf\_parameters \\
psf\_alpha4 & -5.0 & 5.0 & psf\_parameters \\
psf\_beta4 & -5.0 & 5.0 & psf\_parameters \\
\bottomrule
\end{tabular}
\caption{Parameter bounds for 25 dimensional \texttt{hsc-y3-shear} problem.}
\label{tab:hsc}
\end{table}

\subsection{Distribution of Values}
\label{sec:boxes}

\begin{figure*}
    \centering
    \begin{tabular}{|c|}
        \hline
        \textbf{Batch Size 1}\\
        \includegraphics[width=0.27\linewidth]{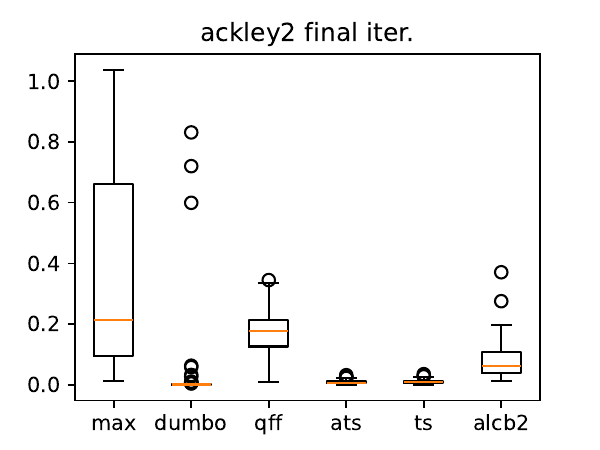}
        \includegraphics[width=0.27\linewidth]{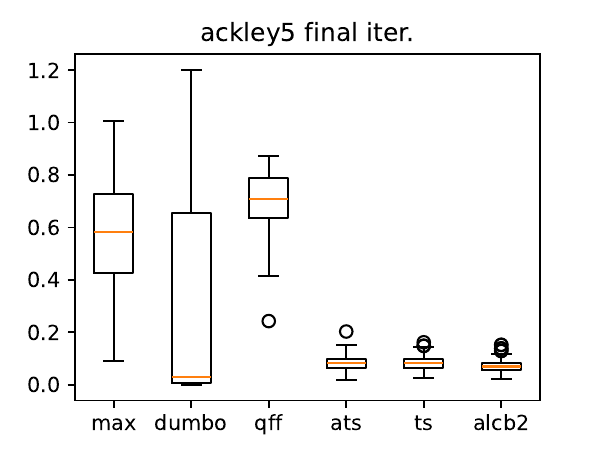}
        \includegraphics[width=0.27\linewidth]{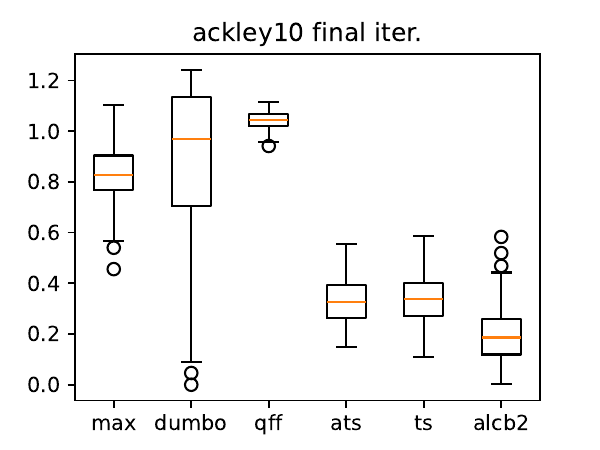}\\
    
        \includegraphics[width=0.27\linewidth]{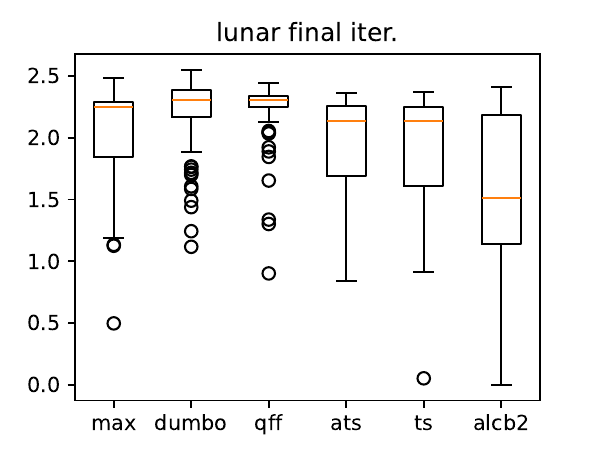}
        \includegraphics[width=0.27\linewidth]{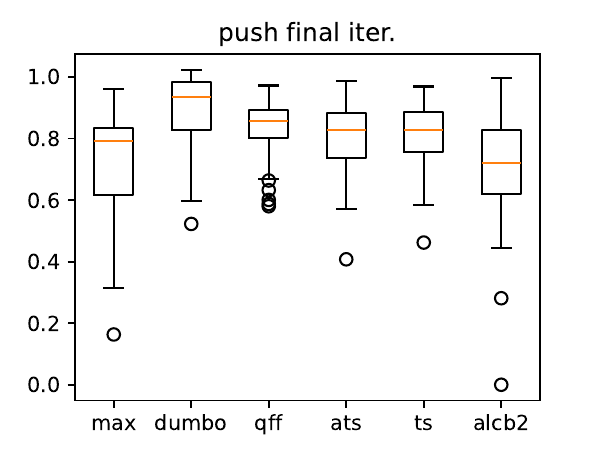}
        \includegraphics[width=0.27\linewidth]{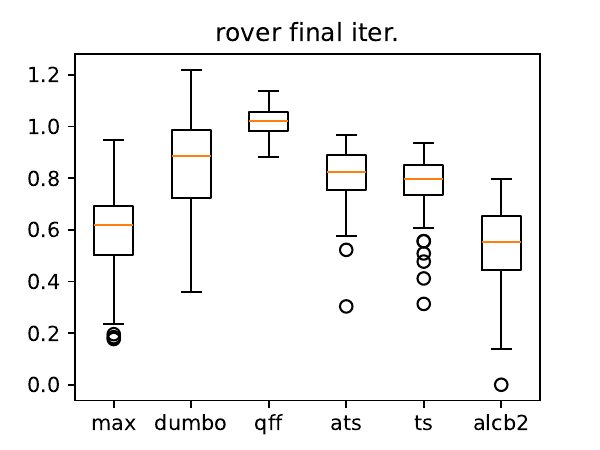}\\
        
        \includegraphics[width=0.27\linewidth]{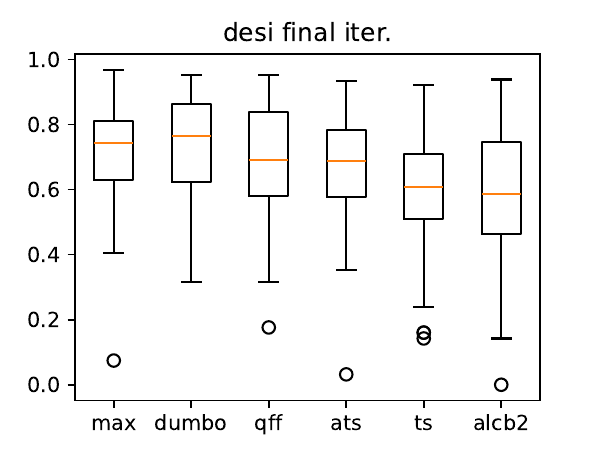}
        \includegraphics[width=0.27\linewidth]{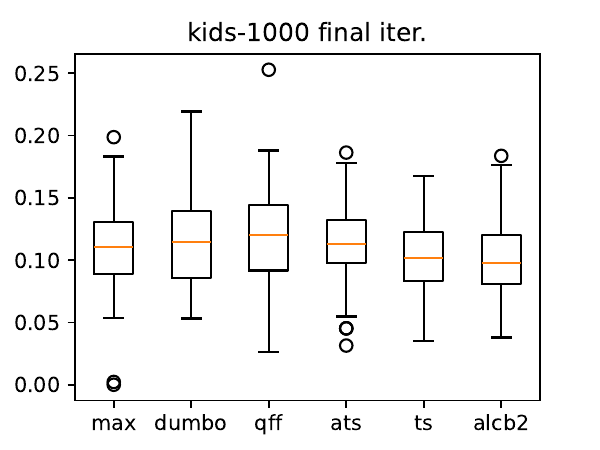}
        \includegraphics[width=0.27\linewidth]{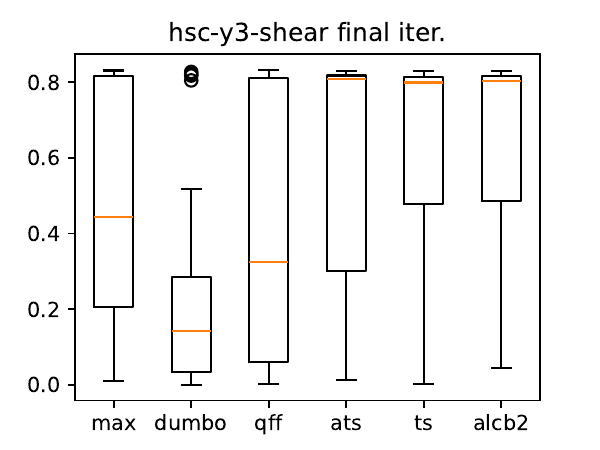}\\
        \textbf{Batch Size 10}\\
        \includegraphics[width=0.27\linewidth]{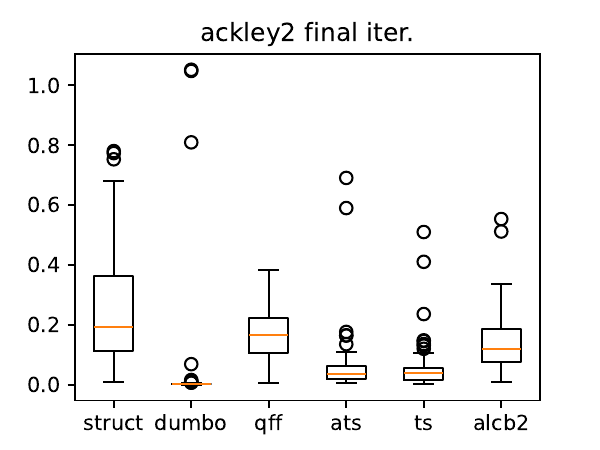}
        \includegraphics[width=0.27\linewidth]{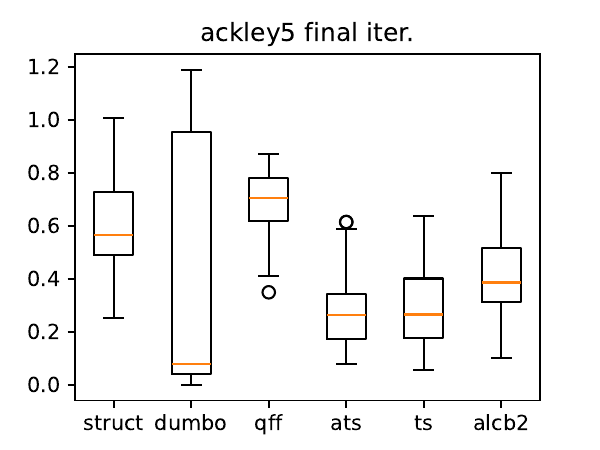}
        \includegraphics[width=0.27\linewidth]{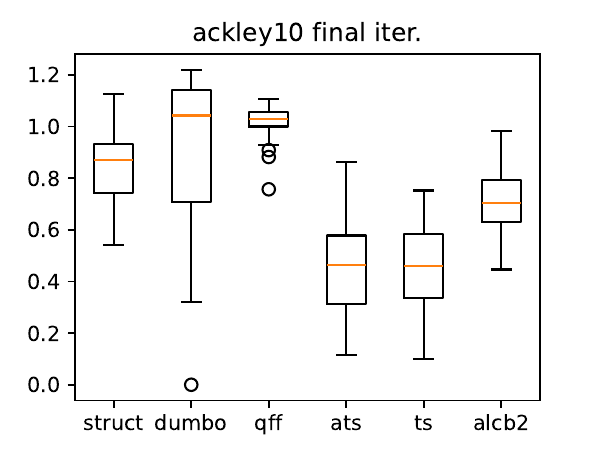}\\
    
        \includegraphics[width=0.27\linewidth]{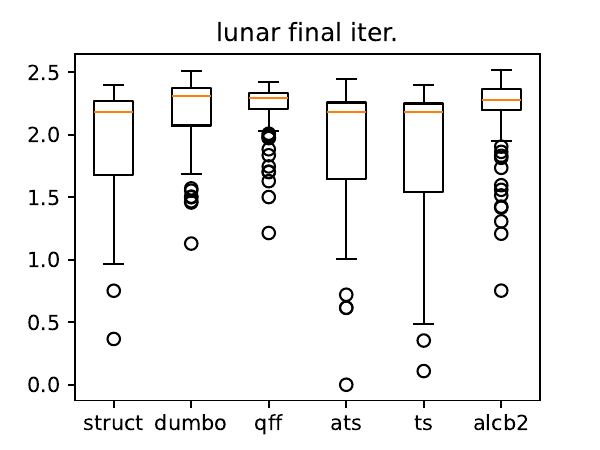}
        \includegraphics[width=0.27\linewidth]{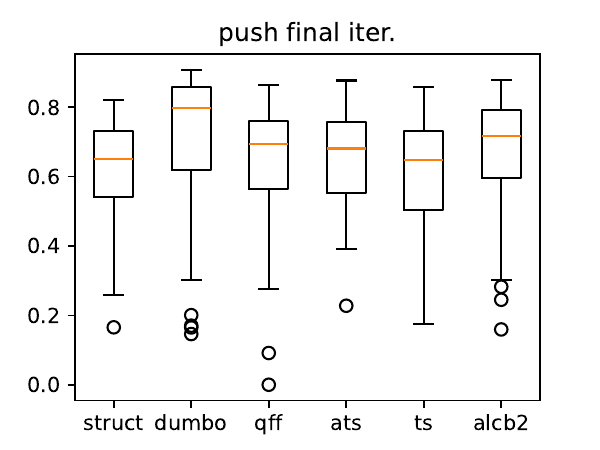}
        \includegraphics[width=0.27\linewidth]{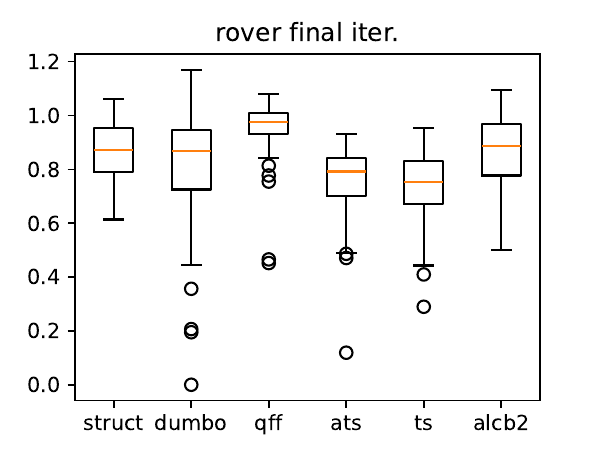}\\
        
        \includegraphics[width=0.27\linewidth]{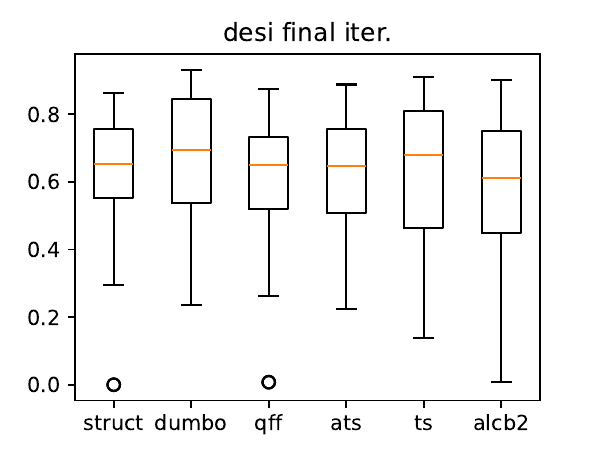}
        \includegraphics[width=0.27\linewidth]{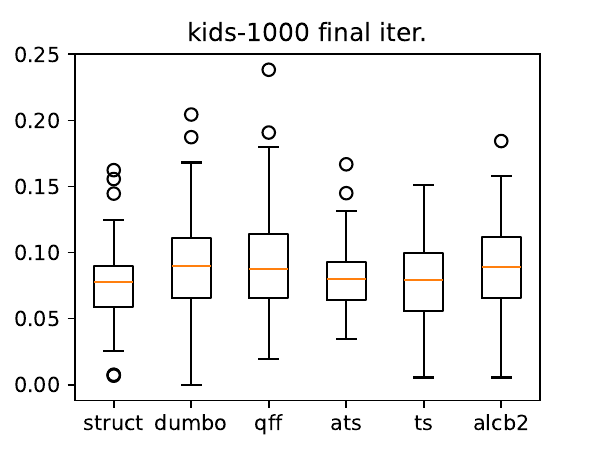}
        \includegraphics[width=0.27\linewidth]{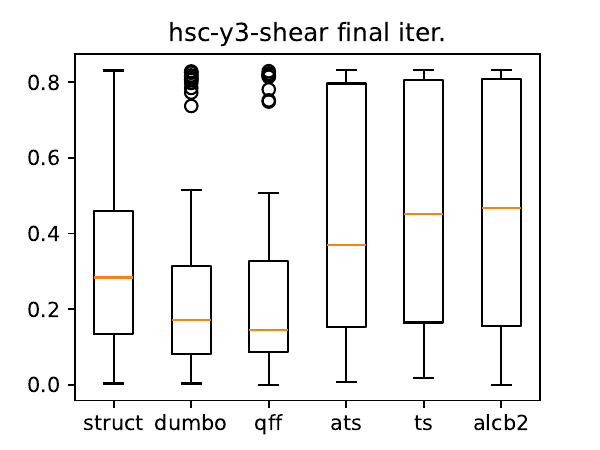}\\
        \hline
    \end{tabular}
    
    \caption{Distribution of log10 optimality gaps for all methods.}
    \label{fig:most_boxes}
\end{figure*}

\begin{figure*}
    \centering
    \begin{tabular}{|c|}
        \hline
        \textbf{Batch Size 1}\\
        \includegraphics[width=0.29\linewidth]{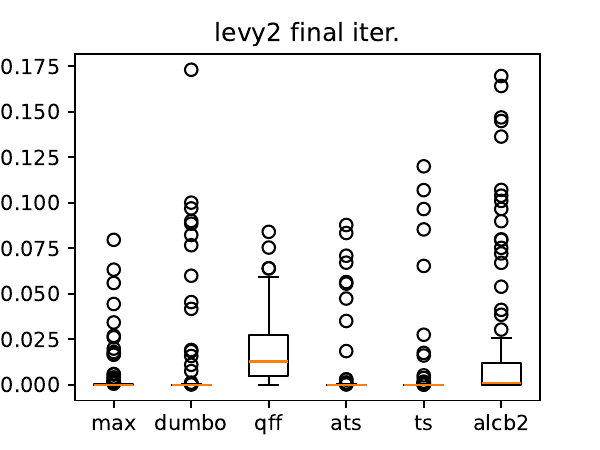}
        \includegraphics[width=0.29\linewidth]{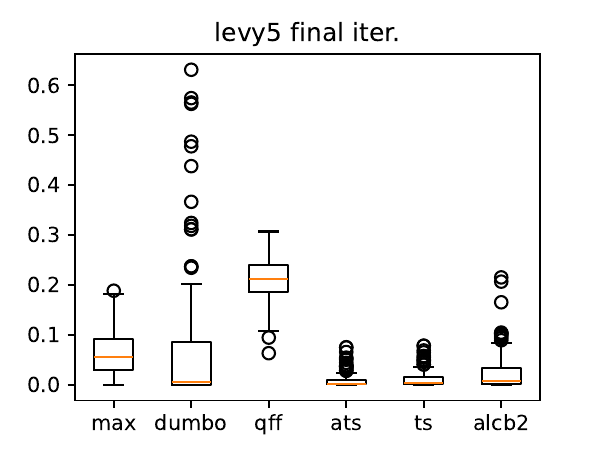}
        \includegraphics[width=0.29\linewidth]{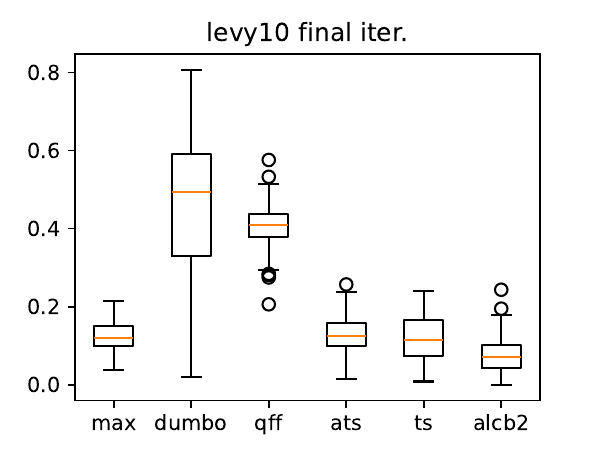}\\
        
        \includegraphics[width=0.29\linewidth]{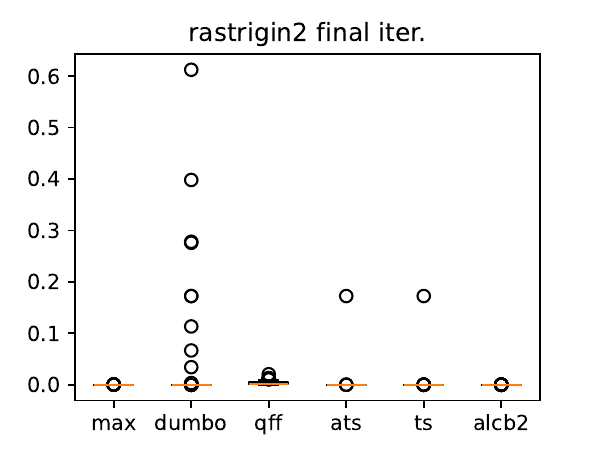}
        \includegraphics[width=0.29\linewidth]{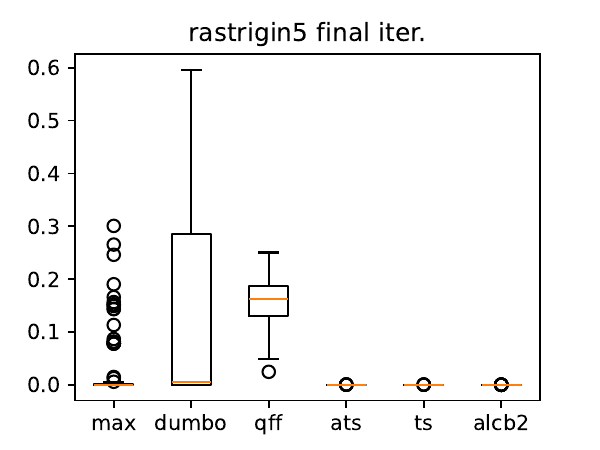}
        \includegraphics[width=0.29\linewidth]{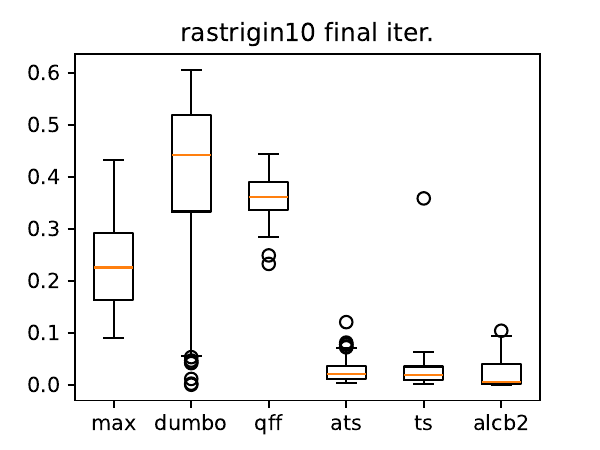}\\

        \textbf{Batch Size 10}\\
        \includegraphics[width=0.29\linewidth]{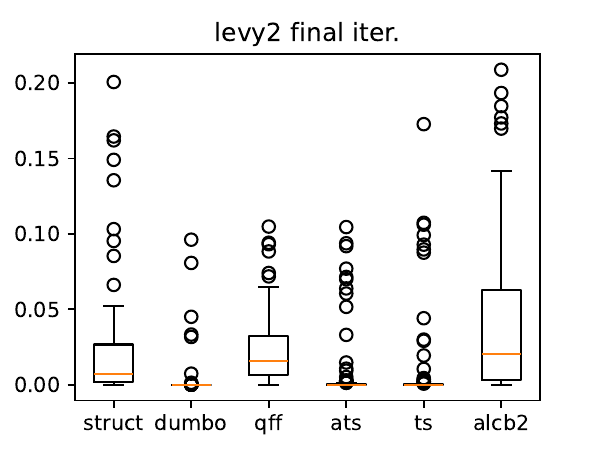}
        \includegraphics[width=0.29\linewidth]{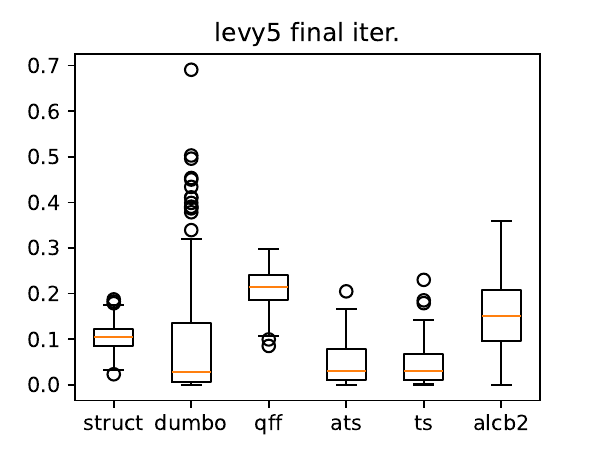}
        \includegraphics[width=0.29\linewidth]{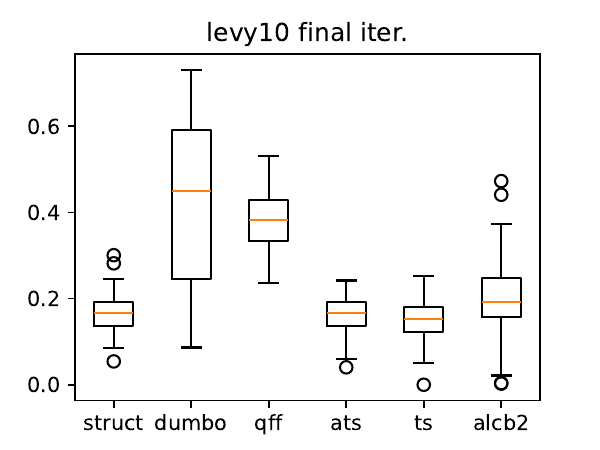}\\
        
        \includegraphics[width=0.29\linewidth]{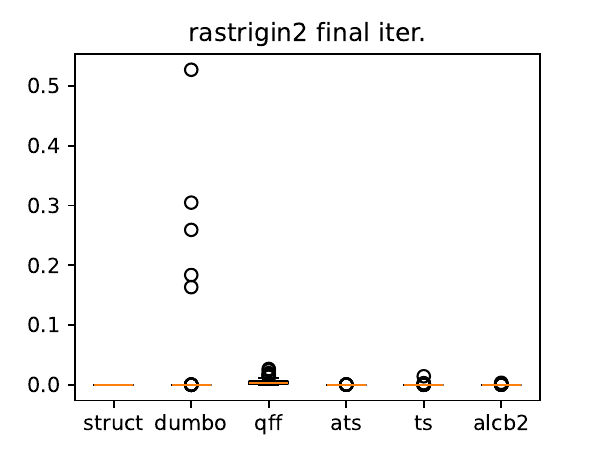}
        \includegraphics[width=0.29\linewidth]{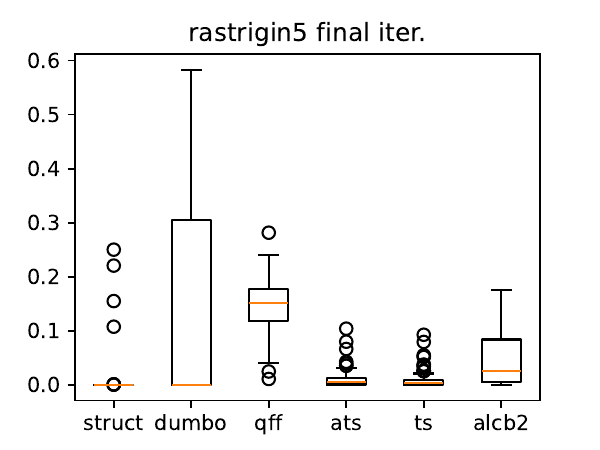}
        \includegraphics[width=0.29\linewidth]{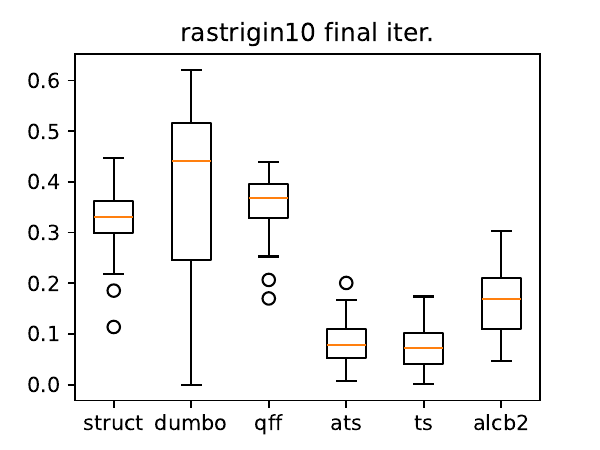}\\
    
        \hline
    \end{tabular}
    
    \caption{Distribution of log10 optimality gaps for all methods.}
    \label{fig:other_boxes}
\end{figure*}

Figures \ref{fig:most_boxes} and \ref{fig:other_boxes} give the distribution of best observed function values for all methods at the final (100th) iteration.
There are a few nuances that our median analysis in the main text.
First, though \texttt{dumbo} does very well on the Ackley problem in 2 and 5 dimensions in terms of median performance, it has among the worst worst-case performance, both in the batch size 1 and 10 cases.
It has similarly bad worst-case performance on the 5 and 10 dimensional Levy and Rastrigin problems.
By contrast, on the Rover problem with a batch size of 10, \texttt{dumbo} actually achieves the best single BOV across any run despite having a median performance inferior to \texttt{ts} and \texttt{ats}.
Finally, we see that the distribution on the Lunar problem is very variable, and medians do not do a great job of capturing the whole distribution, particularly in the batch size 10 case.

\subsection{Impact of Bilateral Uncertainty}
\label{sec:bilateral}

\begin{figure}
    \centering
    \includegraphics[width=0.3\linewidth]{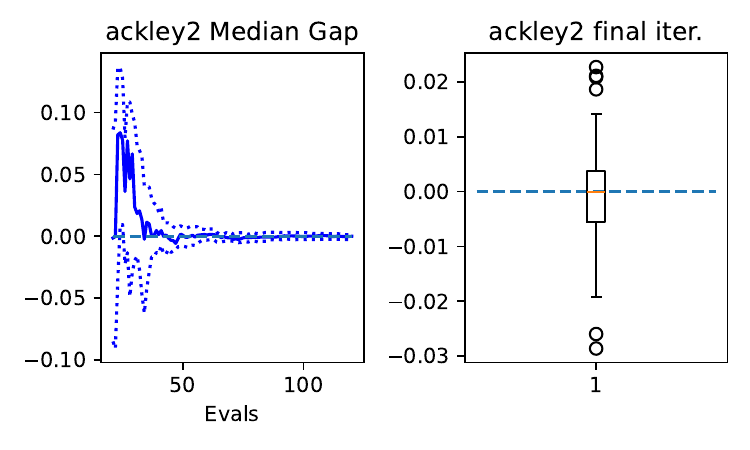}
    \includegraphics[width=0.3\linewidth]{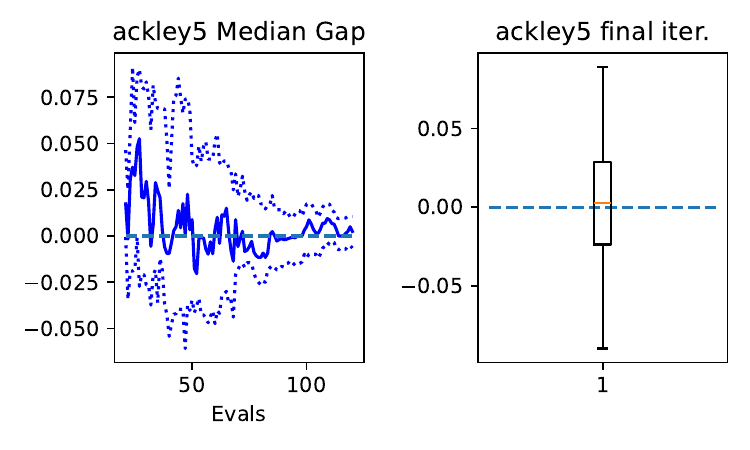}
    \includegraphics[width=0.3\linewidth]{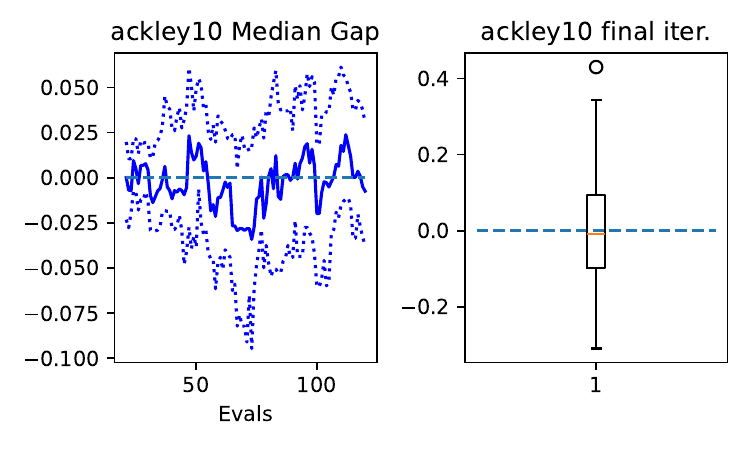}
    
    \includegraphics[width=0.3\linewidth]{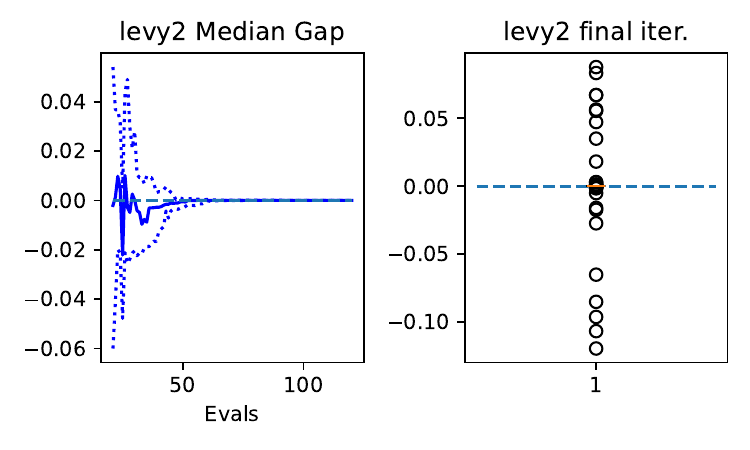}
    \includegraphics[width=0.3\linewidth]{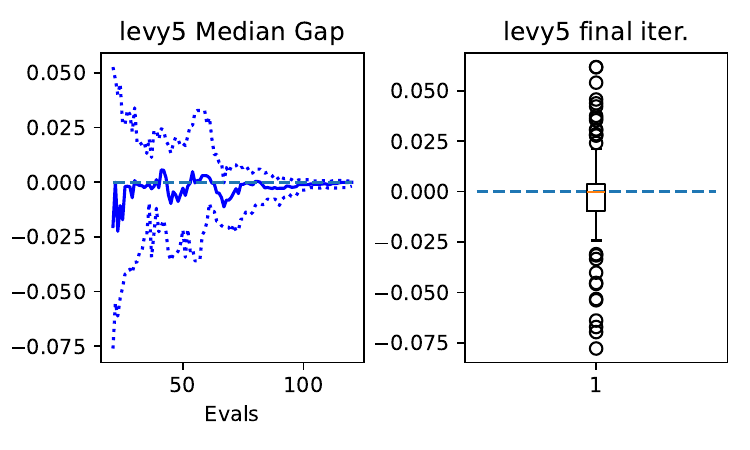}
    \includegraphics[width=0.3\linewidth]{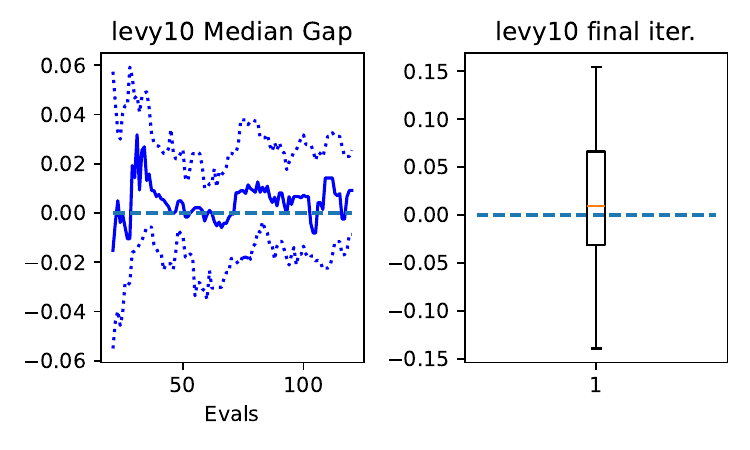}
    
    \includegraphics[width=0.3\linewidth]{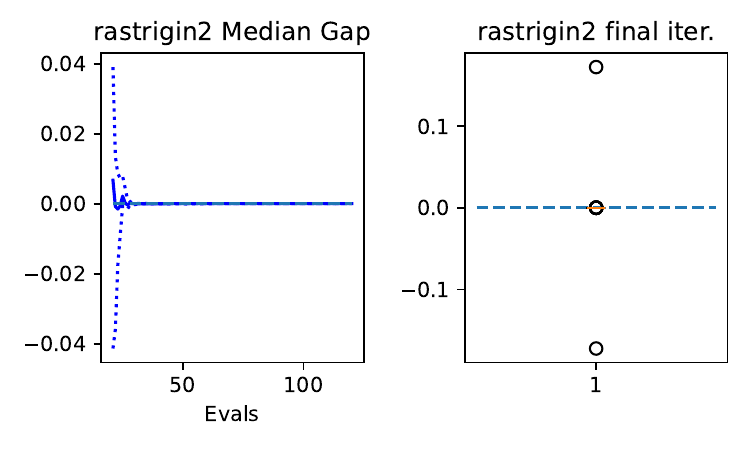}
    \includegraphics[width=0.3\linewidth]{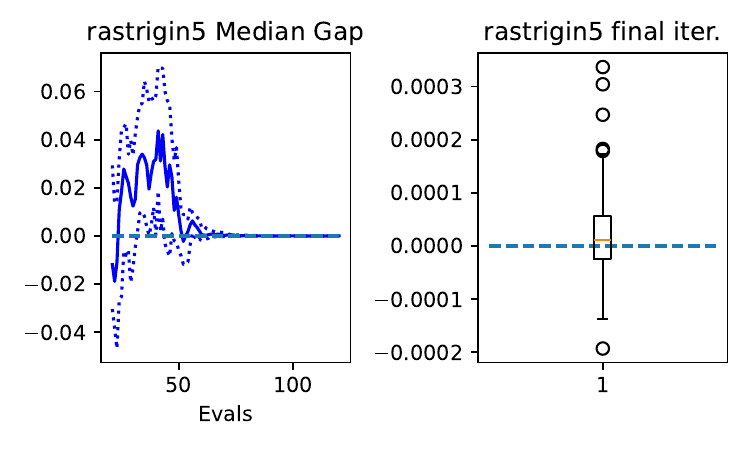}
    \includegraphics[width=0.3\linewidth]{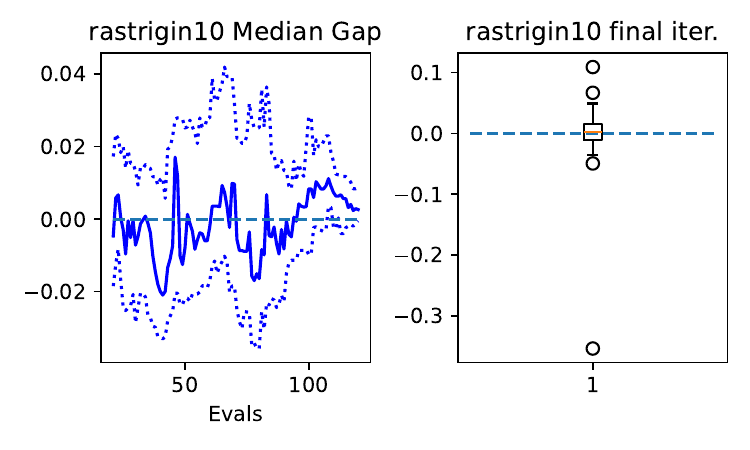}
    
    \includegraphics[width=0.3\linewidth]{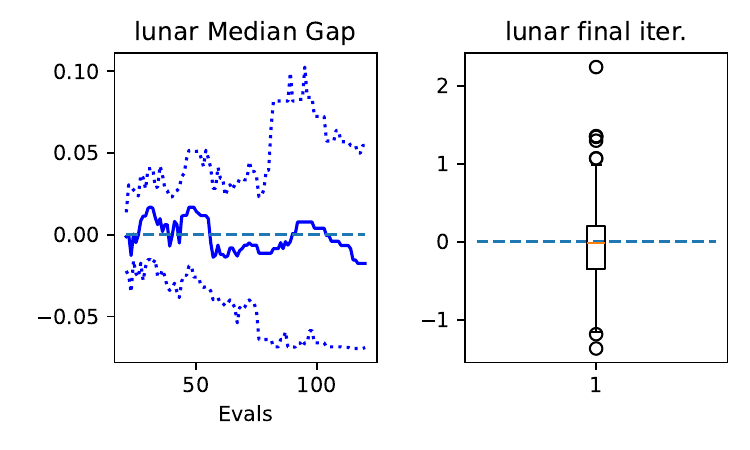}
    \includegraphics[width=0.3\linewidth]{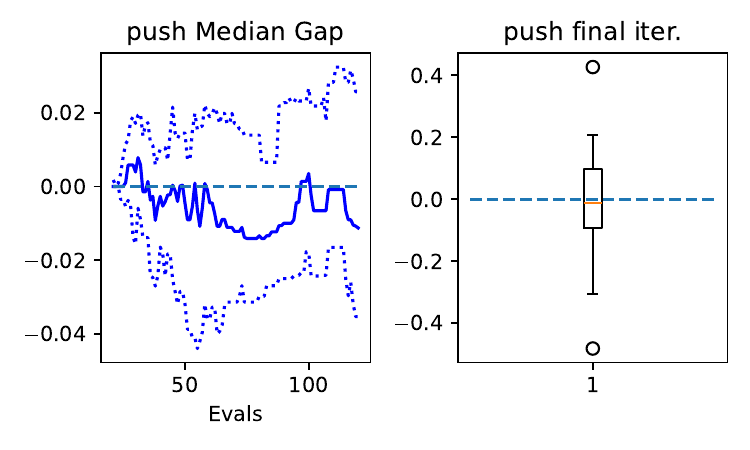}
    \includegraphics[width=0.3\linewidth]{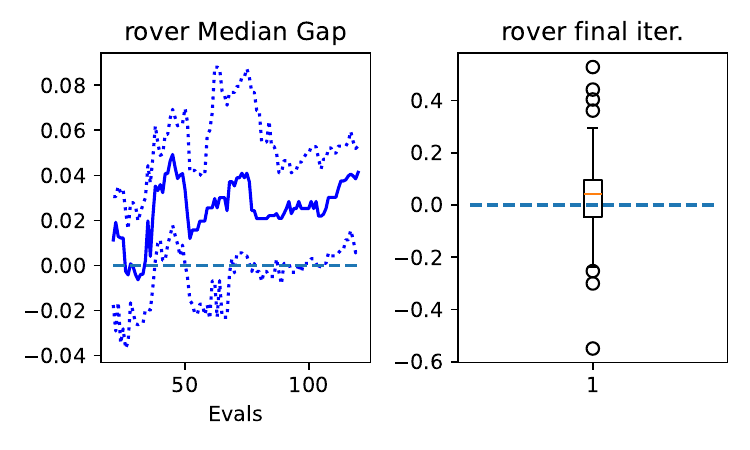}
    
    \includegraphics[width=0.3\linewidth]{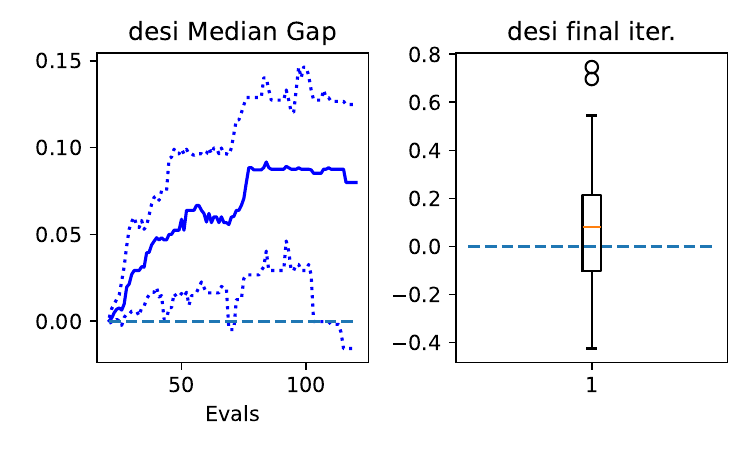}
    \includegraphics[width=0.3\linewidth]{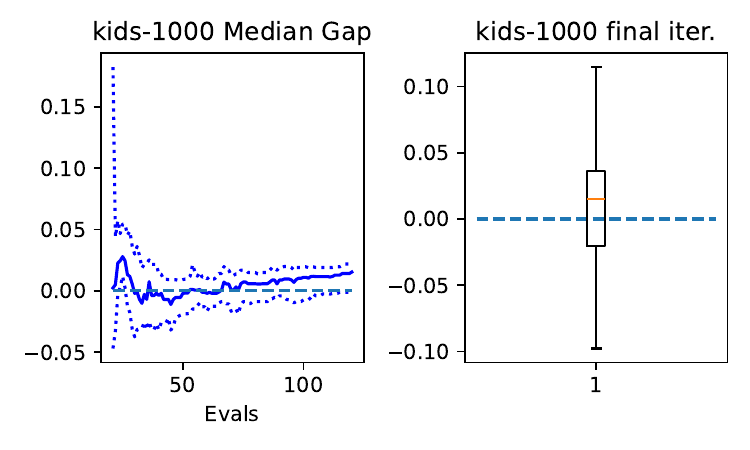}
    \includegraphics[width=0.3\linewidth]{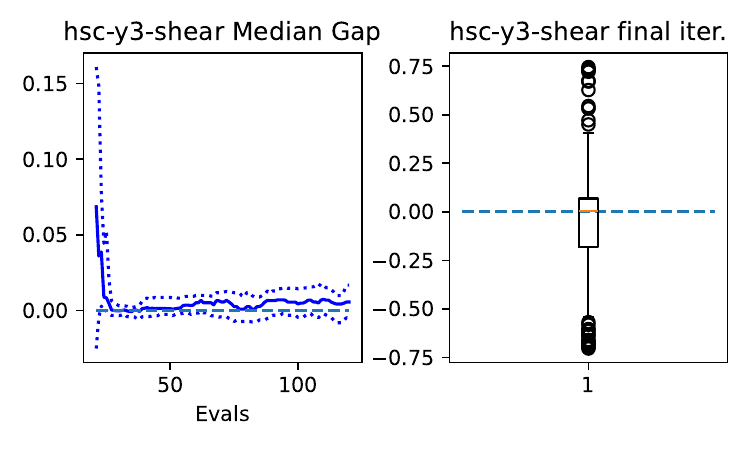}
    \caption{Gap between \texttt{ts} and \texttt{ats} performance for Batch Size 1. In left pane of each set, solid line gives median difference and dotted lines give pointwise 95\% asymptotic confidence interval. Right pane gives a histogram of all 100 runs.}
    \label{fig:diff_bs1}
\end{figure}

\begin{figure}
    \centering
    \includegraphics[width=0.3\linewidth]{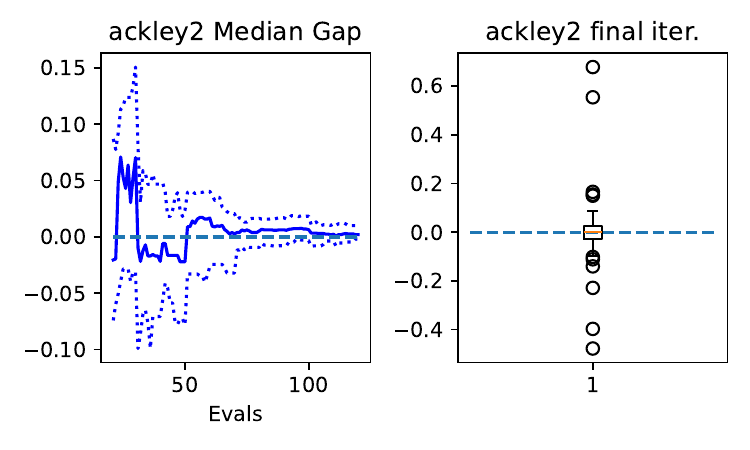}
    \includegraphics[width=0.3\linewidth]{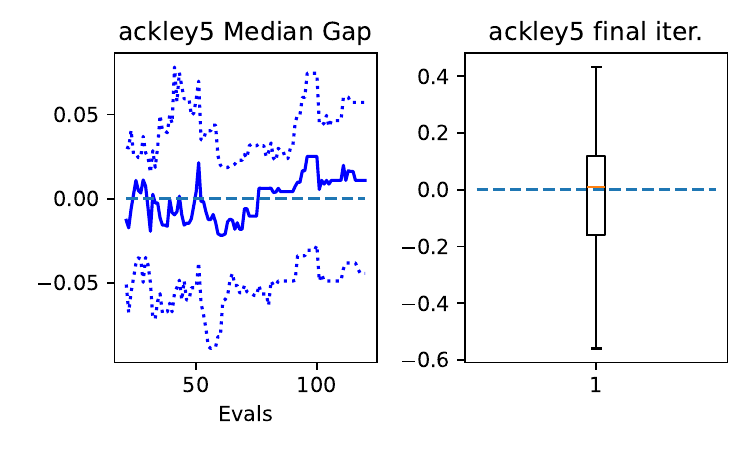}
    \includegraphics[width=0.3\linewidth]{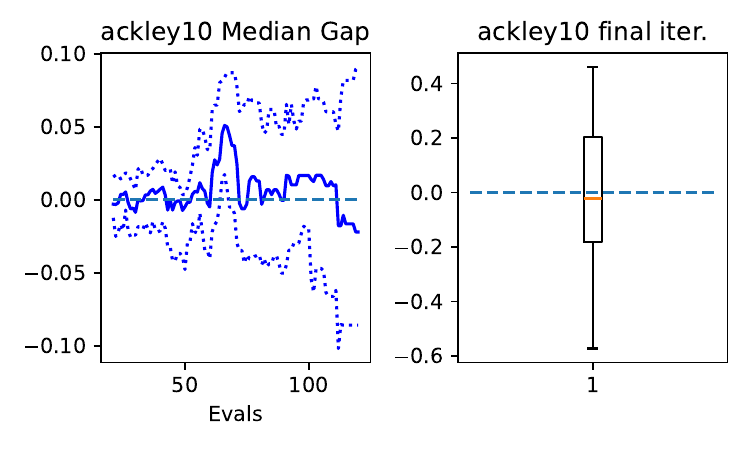}
    
    \includegraphics[width=0.3\linewidth]{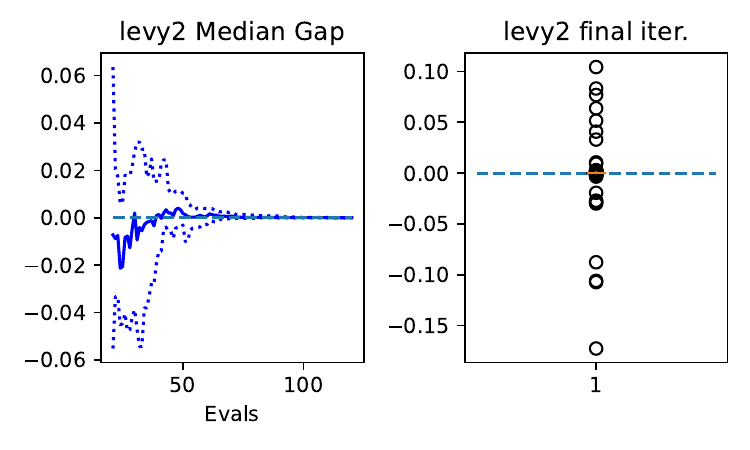}
    \includegraphics[width=0.3\linewidth]{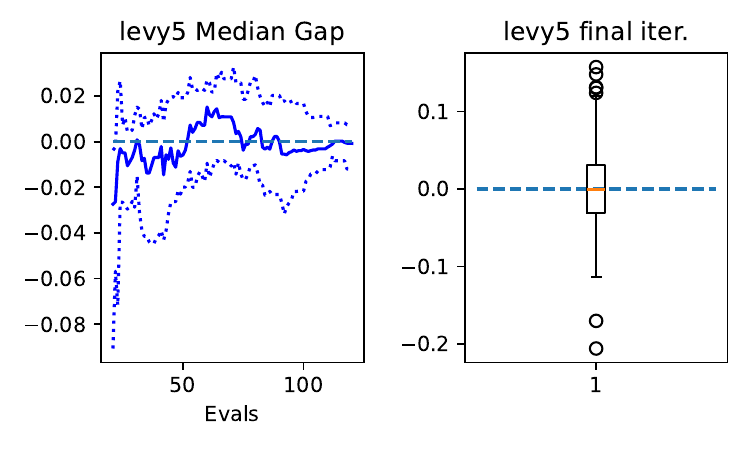}
    \includegraphics[width=0.3\linewidth]{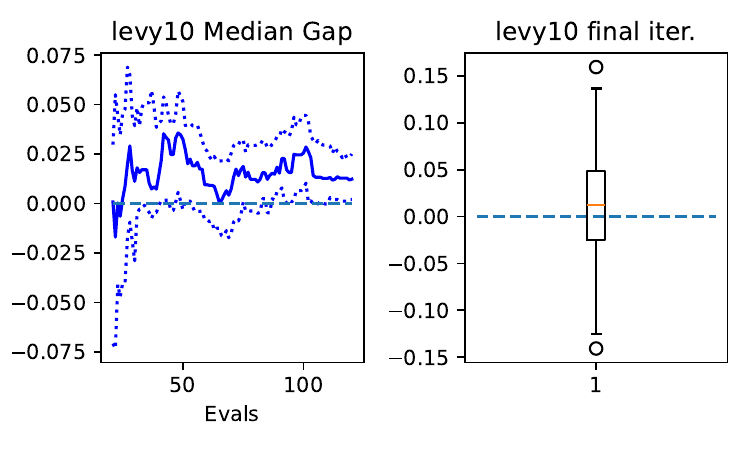}
    
    \includegraphics[width=0.3\linewidth]{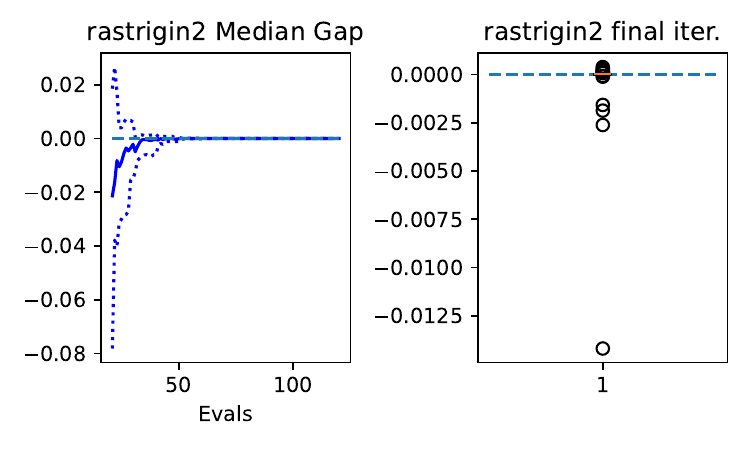}
    \includegraphics[width=0.3\linewidth]{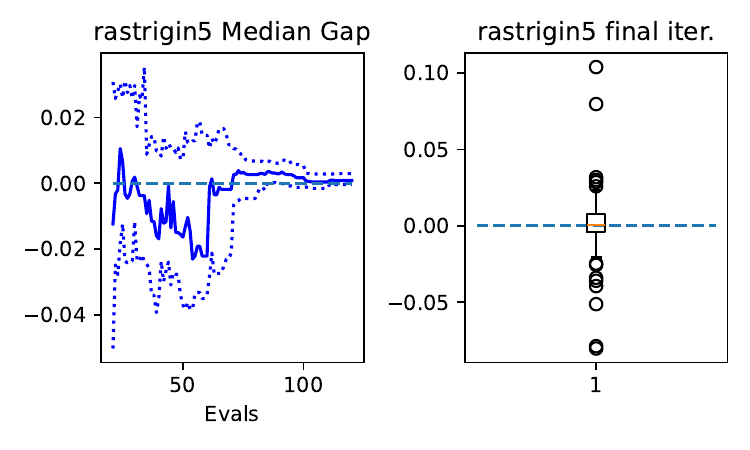}
    \includegraphics[width=0.3\linewidth]{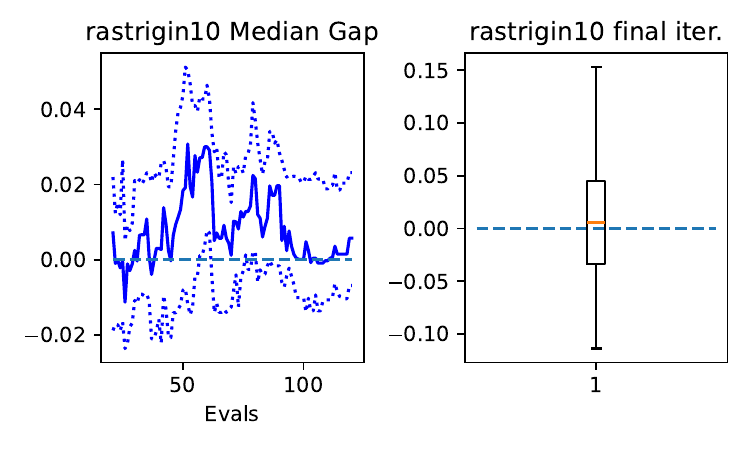}
    
    \includegraphics[width=0.3\linewidth]{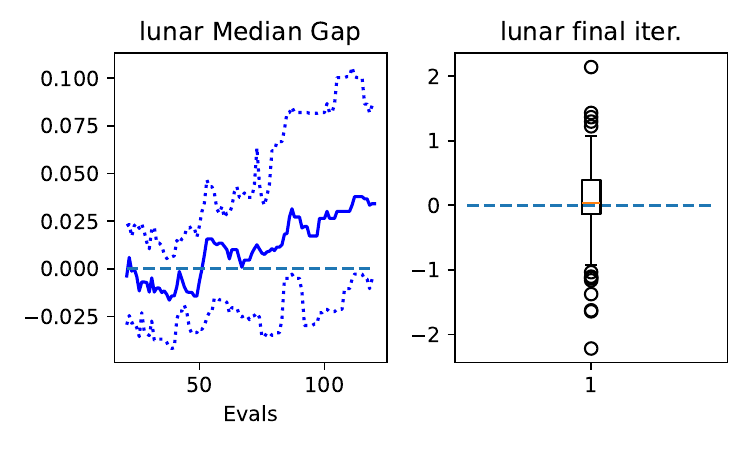}
    \includegraphics[width=0.3\linewidth]{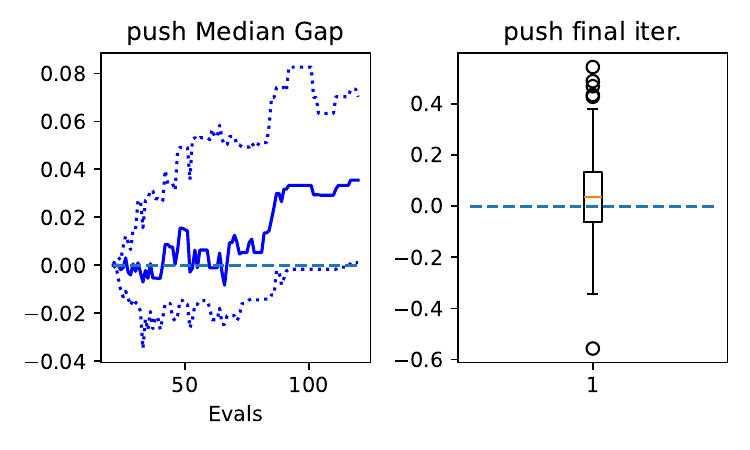}
    \includegraphics[width=0.3\linewidth]{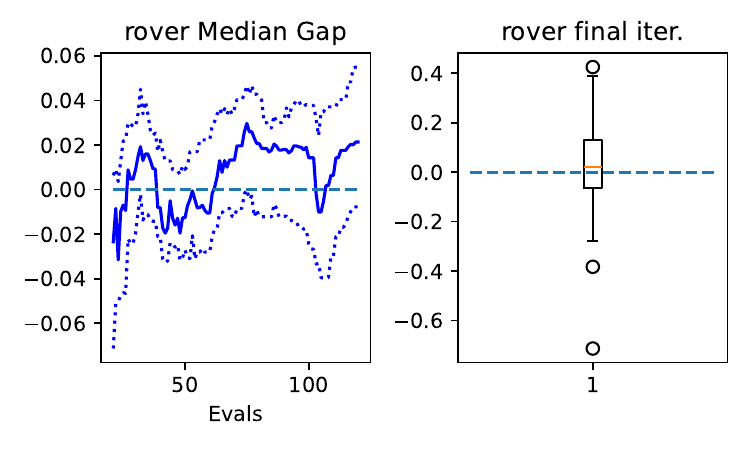}
    
    \includegraphics[width=0.3\linewidth]{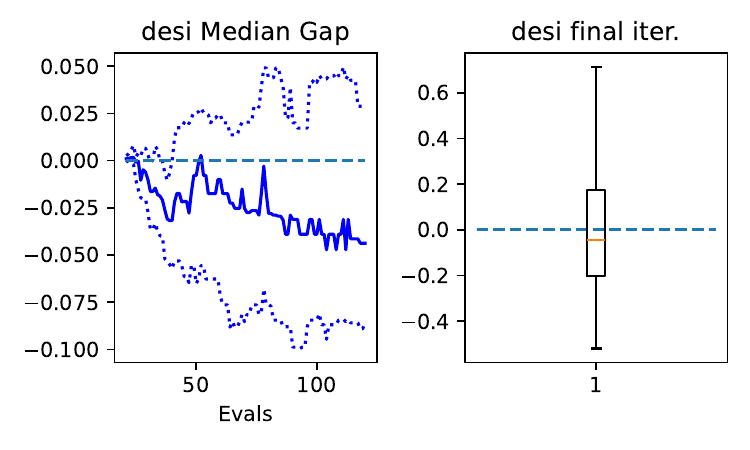}
    \includegraphics[width=0.3\linewidth]{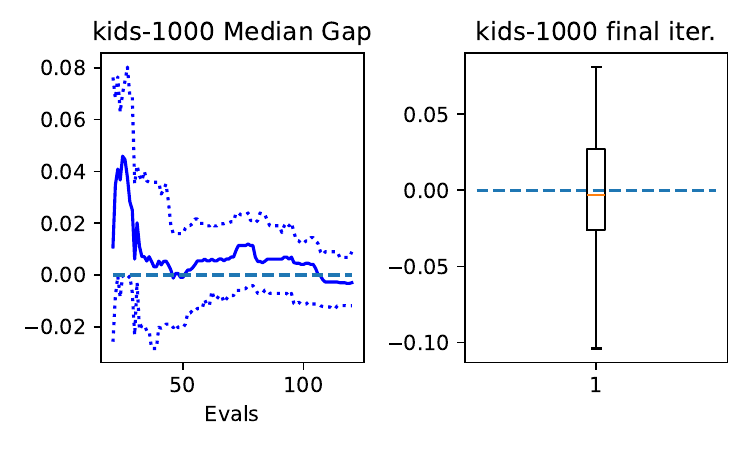}
    \includegraphics[width=0.3\linewidth]{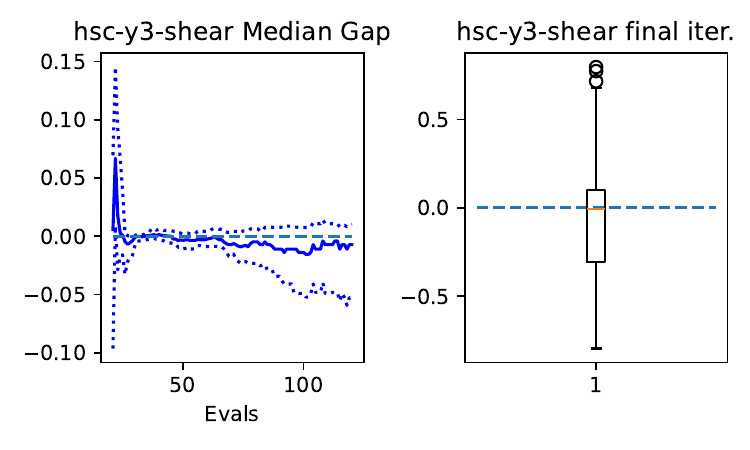}
    \caption{Gap between \texttt{ts} and \texttt{ats} performance for Batch Size 10; see Figure \ref{fig:diff_bs1} caption.}
    \label{fig:diff_bs10}
\end{figure}

Figures \ref{fig:diff_bs1} and \ref{fig:diff_bs10} present our results with a focus on the contrast between the \texttt{ats} and \texttt{ts} methods.
The overall trend is evident: the methods tend not to differ significantly, even when the 95\% interval on the median difference does not cover $0$, suggesting a statistically significant difference, it is typically of a small magnitude.
The boxplots show us that the run-to-run variation invariably outweighs the inter-method variation.
With a batch size of 1 (Figure \ref{fig:diff_bs1}), we find that the \texttt{desi}, \texttt{kids-1000} and \texttt{rover} problems show the greatest advantage for \texttt{ts}.
Intriguingly, the use of nontrivial batches seems to erode this advantage (Figure \ref{fig:diff_bs10}).

\end{document}